\documentclass{article}[11pt]

\usepackage{microtype}
\usepackage{subfigure}
\usepackage{booktabs} %
\usepackage{makecell}
\usepackage{subfigure}
\usepackage{paralist}
\usepackage[usenames,dvipsnames]{xcolor}

\usepackage[pdftex]{graphicx}
\usepackage{tabularx}
\usepackage{lmodern}
\usepackage[colorlinks=true,citecolor=blue]{hyperref}       %

\usepackage{authblk}
\usepackage{natbib}

\newcommand{\newedit}[1]{{#1}}

\usepackage{caption}
\usepackage{wrapfig}

\usepackage{amsmath,amsfonts,amsthm,bm}

\def\1{\bm{1}}

\DeclareMathAlphabet{\mathsfit}{\encodingdefault}{\sfdefault}{m}{sl}
\SetMathAlphabet{\mathsfit}{bold}{\encodingdefault}{\sfdefault}{bx}{n}

\DeclareMathOperator*{\argmax}{arg\,max}

\newcommand{\XCal}{\mathscr{X}}
\newcommand{\YCal}{\mathscr{Y}}

\usepackage[utf8]{inputenc} %
\usepackage[T1]{fontenc}    %
\usepackage{url}            %
\usepackage{booktabs}       %
\usepackage{amsfonts}       %
\usepackage{nicefrac}       %
\usepackage{microtype}      %

\usepackage{multirow}
\usepackage{float}
\usepackage{graphicx}
\usepackage[export]{adjustbox}
\usepackage[inline]{enumitem}

\usepackage{amsmath,amssymb,amsthm}
\usepackage[mathscr]{eucal}
\usepackage{stmaryrd}
\usepackage{accents}
\usepackage{upgreek}

\usepackage{booktabs,colortbl,multirow}

\usepackage{soul}
\usepackage{array}

\newcommand{\helps}[1]{\cellcolor{blue!25}{#1}}
\newcommand{\harms}[1]{\cellcolor{red!50}{#1}}
\newcommand{\neutral}[1]{\cellcolor{green!25}{#1}}

\graphicspath{{figs//}}

\allowdisplaybreaks

\newcommand\numberthis{\addtocounter{equation}{1}\tag{\theequation}}

\usepackage[normalem]{ulem}

\newcommand{\defEq}{\stackrel{.}{=}}

\renewcommand{\Pr}{\mathbb{P}}

\newcommand{\Real}{\mathbb{R}}

\theoremstyle{definition}

\newcommand{\answerYes}[1][]{\textcolor{blue}{[Yes] #1}}
\newcommand{\answerNo}[1][]{\textcolor{orange}{[No] #1}}
\newcommand{\answerNA}[1][]{\textcolor{gray}{[N/A] #1}}
\newcommand{\answerTODO}[1][]{\textcolor{red}{\bf [TODO]}}

\usepackage[left=1.25in, top=1in, bottom=1in, right=1.25in]{geometry}
\title{Teacher's pet: understanding and mitigating biases in distillation}
\author{Michal Lukasik, Srinadh Bhojanapalli, Aditya Krishna Menon and Sanjiv Kumar \\
Google Research, New York \\
\texttt{\{mlukasik,bsrinadh,adityakmenon,sanjivk\}@google.com}}

\begin{document}

\maketitle

\begin{abstract}
    Knowledge distillation is widely used as a means of improving the performance of a relatively simple ``student'' model using the predictions from a complex ``teacher'' model.
Several works have shown that distillation significantly boosts the student's \emph{overall} performance;
however, are these gains uniform across all data subgroups?
In this paper, we
show that
distillation can \emph{harm} performance on 
certain subgroups,
{e.g., classes with few associated samples}.
We trace this behaviour to errors made by the teacher distribution being transferred to and \emph{amplified} by the student model.
To mitigate this problem,
we present techniques 
which soften the teacher influence for subgroups where it is less reliable.
Experiments on several image classification benchmarks show that these modifications of distillation maintain boost in overall accuracy,
while additionally ensuring improvement in subgroup performance.

\end{abstract}

\section{Introduction}
Knowledge distillation is a 
technique for improving the performance of a ``student'' model using the predictions from a  ``teacher'' model.
At its core,
distillation involves
replacing the one-hot training labels with the teacher's predicted distribution over labels.
Empirically,
distillation has proven successful 
as a means of
model compression~\citep{Bucilua:2006,Hinton:2015},
improving the performance of a fixed model architecture~\citep{Anil:2018,Furlanello:2018},
and 
semi-supervised learning~\citep{Radosavovic:2018}.
Theoretically, considerable recent effort~\citep{Lopez-Paz:2016,Mobahi:2020,Tang:2020,Menon:2020,Zhang:2020b,Ji:2020,AllenZhu:2020,Zhou:2021,Dao:2021} has 
focused on understanding how distillation affects learning.
Put together, both strands of work further the understanding of when and why distillation helps.

In this paper, we are similarly motivated to better understand the mechanics of distillation,
but pose a slightly different question:
does distillation help \emph{all} data subgroups uniformly?
Or, do its overall gains come at the expense of \emph{degradation} of performance on certain subgroups?
To our knowledge,
there has been no systematic study (empirical or otherwise) of this question.
This consideration is topical given the study of \emph{fairness} of machine learning algorithms on under-represented subgroups~\citep{Hardt:2016,Buolamwini:2018,Chzhen:2019,Sagawa:2020}.

\begin{figure}[!t]
    \centering
    \resizebox{0.85\linewidth}{!}{
     \subfigure[\scriptsize CIFAR-100-LT]{
    \includegraphics[scale=0.15]{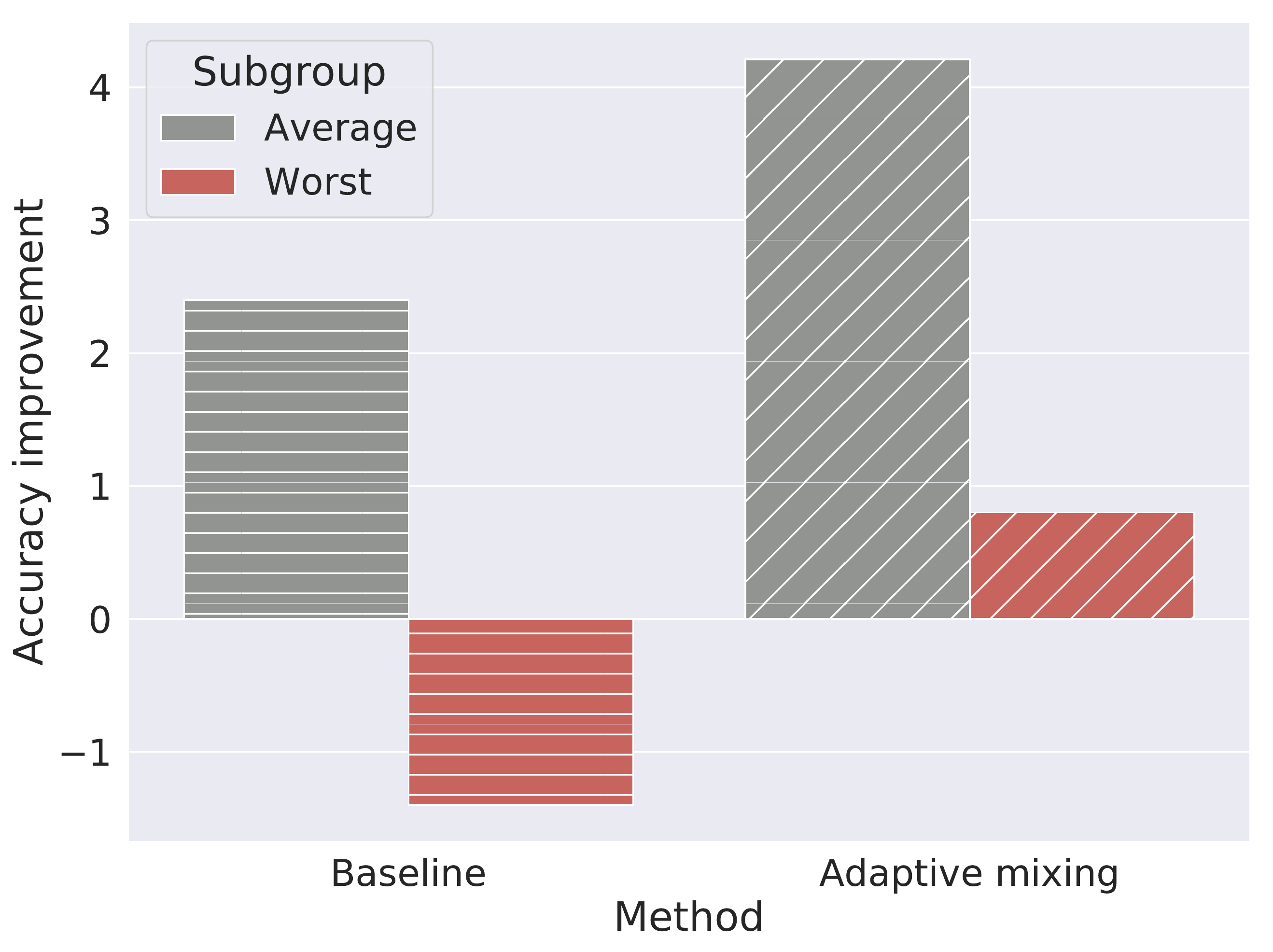}
    }
    \hfill
    \subfigure[\scriptsize Imagenet]{
    \includegraphics[scale=0.15]{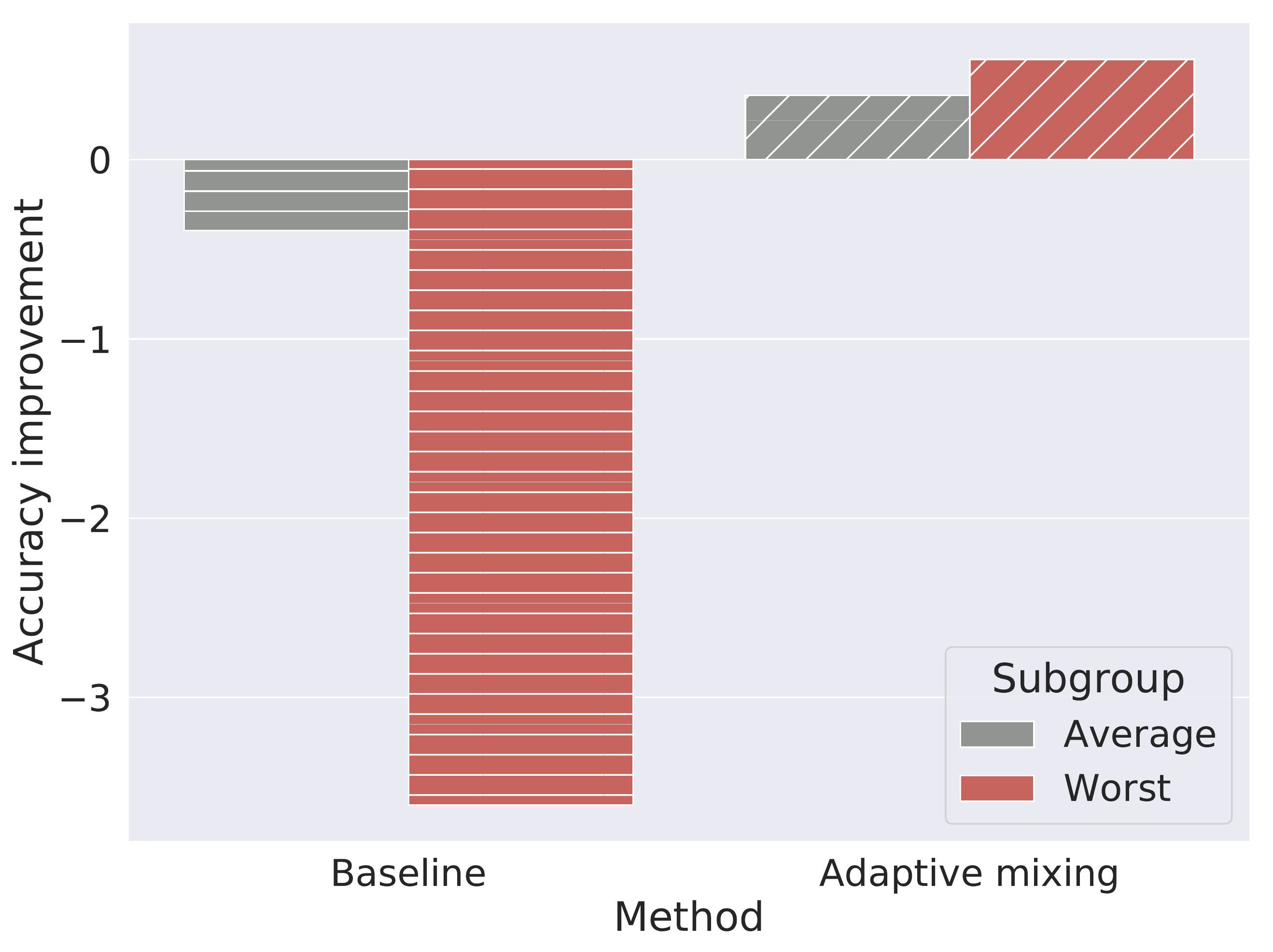}
    }
    } 
    \caption{Illustration of the potential deleterious effects of distillation on data subgroups.
    We train a 
    ResNet-56 teacher on \textbf{CIFAR-100-LT}, a long-tailed version of CIFAR-100~\citep{Cui:2019,Cao:2019} where some labels have only a few associated samples,
    and a
    ResNet-50 teacher on \textbf{ImageNet}.
    For each dataset, we self-distill to a student ResNet 
    of the same depth.
    On CIFAR-100-LT, 
    as is often observed,
    distillation reasonably helps the \emph{overall} accuracy over one-hot student training ($\sim$1\% absolute).
    However, such gains come at significant cost on subgroups defined by the individual classes:
    on the ten classes with lowest label frequency,
    distillation \emph{harms} performance by $\sim$2\% absolute.
    Similarly, on Imagenet, %
    distillation {harms} 
    the average accuracy of the worst-10 classes
    by $\sim$3\% absolute.
    Our proposed techniques (\S\ref{sec:methods}) can roughly preserve the overall accuracy, 
    while boosting subgroup performance.
    }
    \label{fig:motivating_example}
\end{figure}

Our first finding is that
even in standard settings
---
e.g., on image classification benchmarks such as CIFAR
---
distillation can disproportionately %
\emph{harm} performance on subgroups defined by the individual classes (see Figure~\ref{fig:motivating_example}).
To discern the source of this behaviour,
we ablate the teacher and student architectures (\S\ref{sec:ablation-architecture}),
dataset complexity (\S\ref{sec:ablation-dataset}),
and label frequencies (\S\ref{sec:ablation-imbalance}).
These point to the potential harms of distillation 
when
the teacher makes
\emph{confident mispredictions} on a subgroup.

Having identified a potential limitation of distillation, we then study how to remedy it.
To this end,
we present 
{two simple techniques
which apply
per-subgroup mixing weights between the teacher and one-hot labels,
and
per-subgroup margins respectively (\S\ref{sec:methods})}.
Intuitively, these seek to limit the influence of teacher predictions on subgroups it models poorly.
Experiments on image classification benchmarks show that 
in many cases,
these modifications maintain boost in overall accuracy,
while ensuring a more equitable improvement across subgroups.
In sum, our contributions are:
\begin{enumerate}[itemsep=0pt,topsep=0pt,leftmargin=16pt,label=(\roman*)]
    \item we identify 
    a potential issue with 
    distillation,
    namely, that its improvements in overall accuracy
    may come at the expense of harming accuracy on 
    certain subgroups (\S\ref{sec:distillation-harms-hard})
    
    \item 
    we ablate potential sources for the above phenomenon (\S\ref{sec:ablation-architecture}, \S\ref{sec:ablation-dataset}, \S\ref{sec:ablation-imbalance}),
    and in the process identify certain characteristics of data (e.g., skewed label distributions) where it can manifest;
    
    \item we propose 
    {two simple modifications} of distillation that mitigate this problem, 
    {based on applying 
    per-subgroup
    mixing weights
    and margins} (\S\ref{sec:methods});
    these are shown to perform well empirically (\S\ref{sec:experiments}).
    
\end{enumerate}

\section{Background and related work}

\textbf{Knowledge distillation}.
Consider a multi-class classification problem over instances $\XCal$ and labels $\YCal = [L] \defEq \{ 1, \ldots, L \}$.
Given a training set $S = \{ ( x_n, y_n ) \}_{n = 1}^N$ drawn from some distribution $\Pr$,
we seek
a classifier $h \colon \XCal \to \YCal$ 
that minimises
the \emph{misclassification error}
\begin{equation}
    \label{eqn:misclassification-error}
    {E}_{\mathrm{avg}}( h ) \defEq \Pr( h( x ) \neq y ).
\end{equation}
In practice, one may learn \emph{logits} $f \colon \XCal \to \Real^L$ to minimise
$ \hat{R}( f ) = \frac{1}{N} \sum_{n = 1}^{N} \ell( y_n, f( x_n ) ), $
where $\ell$ is a loss function such as the softmax cross-entropy,
$\ell( y, f( x ) ) = -f_y( x ) + \log\left[ \sum_{y' \in [L]} e^{f_{y'}( x )} \right]$.
One may then classify the sample via $h( x ) = \argmax_{y \in [L]} f_y( x )$.

Knowledge distillation~\citep{Bucilua:2006,Hinton:2015}
employs
the logits $f^{\mathrm{t}} \colon \XCal \to \Real^L$ of
a %
``teacher'' model to train a 
``student'' model.
The latter learns logits $f^{\rm s} \colon \XCal \to \Real^L$ to minimise:
\begin{equation}
    \label{eqn:distillation}
    \begin{aligned}
        \hat{R}_{\mathrm{dist}}( f ) = \frac{1}{N} \sum_{n = 1}^{N} \big[ &( 1 - \alpha ) \cdot \ell( y_n, f( x_n ) ) + \alpha \cdot \sum_{y' \in [L]} p^{\mathrm{t}}_{y'}( x_n ) \cdot \ell( y', f( x_n ) ) \big],
    \end{aligned}
\end{equation}
where $\alpha \in [0, 1]$.
Here, one converts the teacher logits to probabilities $p^{\mathrm{t}} \colon \XCal \to \Delta_{L}$
for simplex $\Delta$,
e.g. via a softmax transformation $p^{\mathrm{t}}_{y'}( x ) \propto \exp( f^{\mathrm{t}}_{y'} ( x ) )$.
The second term \emph{smooths} the student labels based on the teacher's confidence that they explain the sample.
The first term includes the original label, so as to prevent incorrect teacher predictions from overwhelming the student.
One further important trick 
is \emph{temperature scaling} of the teacher logits,
so that $p^{\mathrm{t}}_{y'}( x ) \propto \exp( T^{-1} \cdot f^{\mathrm{t}}_{y'} ( x ) )$.
Setting $T \gg 0$ makes $p^{\mathrm{t}}$ more uniform,
and thus guards against overconfident predictions~\citep{Guo:2017}.

\textbf{Average versus subgroup performance}.
The above exposition treats the
misclassification error ${E}_{\mathrm{avg}}( h )$~\eqref{eqn:misclassification-error}
as the fundamental performance measure of interest.
However, 
suppose the 
data
contains
\emph{subgroups} $\mathscr{G} = \{ 1, \ldots, G \}$.
\newedit{Defining the \emph{per-subgroup errors} $\mathrm{err}_g( h ) \defEq \Pr( h( x ) \neq y \mid g )$,}
we have
$ {E}_{\mathrm{avg}}( h ) = \sum_{g \in \mathscr{G}} \Pr( g ) \cdot \mathrm{err}_g( h ), $
which may mask errors on samples with 
$\Pr( g ) \sim 0$~\citep{Sagawa:2020,Sagawa:2020b,Sohoni:2020}.
\newedit{To this end, one may instead measure the \emph{balanced} error~\citep{Menon:2013}
${E}_{\mathrm{bal}}( h ) \defEq \sum_{g \in \mathscr{G}} \frac{1}{|\mathscr{G}|} \cdot \mathrm{err}_g( h )$
which treats the subgroup distribution as uniform,
or the \emph{worst-subgroup} error~\citep{Sagawa:2020,Sagawa:2020b,Sohoni:2020}
${E}_{\mathrm{max}}( h ) \defEq \max_{g \in \mathscr{G}} \mathrm{err}_g( h )$,
which focusses on the worst-performing subgroup.
An intermediary is the average of the $k$ worst-performing subgroups~\citep{Williamson:2019}:
if $\mathrm{err}^{[i]}( h )$ denotes the $i$th largest per-subgroup error,
\begin{equation}
    \label{eqn:top-k}
    {E}_{\mathrm{top-k}}( h ) \defEq \frac{1}{k} \sum_{i = 1}^{k} \mathrm{err}^{[i]}( h ).
\end{equation}}

The definition of subgroups is a domain-specific consideration.
One important special case is where each label defines a subgroup (i.e., $\mathscr{G} = \mathscr{Y}$),
and $\Pr( y )$ is skewed.
In such \emph{long-tail} settings~\citep{Buda:2017,VanHorn:2017},
classifiers with good average performance can
perform poorly  on ``tail'' labels where $\Pr( y ) \sim 0$.

\textbf{Related work}.
There is limited prior study that dissects distillation's overall gains per subgroup.
\citet{Zhao:2020} showed that in incremental learning settings, 
distillation can 
be biased towards recently observed classes.
We show that even in offline settings,
distillation can harm certain classes. %
\newedit{Recently,~\citet{Zhou:2021} studied the 
standard \emph{aggregate}~\eqref{eqn:misclassification-error} performance of distillation,
which was tied to a certain subset of ``regularisation samples''.
By contrast, our primary concern is to understand the \emph{subgroup} performance of distillation.}
Study of the \emph{fairness} of machine learning algorithms on under-represented data subgroups has received recent attention~\citep{Dwork:2012,Hardt:2016,Buolamwini:2018,Chzhen:2019}.
This has prompted dissection of the performance of established techniques,
such as 
dimensionality reduction~\citep{Samadi:2018},
increasing model capacity~\citep{Sagawa:2020},
and selective classification algorithms~\citep{Jones:2021}.
We follow the general spirit of such works,
studying a more delicate setting involving \emph{two} separate models (the student and teacher),
each with their own inductive biases. 
We present more discussion of related directions in \S\ref{sec:conclusion}.

\section{Are distillation's gains uniform?}
\label{sec:analysis}
We demonstrate that
the gains of distillation may \emph{not be uniform across subgroups}:
\newedit{specifically, considering subgroups defined by classes,}
distillation can \emph{harm} the student's performance
on \newedit{the ``hardest'' few classes} (\S\ref{sec:distillation-harms-hard}).
To understand the genesis of this problem,
we perform ablations 
(cf.~Table~\ref{tbl:analysis_summary})
that establish
its existence in settings where there are insufficient samples to model certain classes,
either due to the number of classes being large (\S\ref{sec:ablation-dataset}),
or the class distribution being skewed (\S\ref{sec:ablation-imbalance}).
We then identify that the student may amplify the teacher's errors (\S\ref{sec:why-harm}).
\newedit{Finally, we corroborate these results for a more general notion of subgroup in a fairness dataset (\S\ref{sec:subgroup-justification}).}

\begin{table}[!t]
    \centering
    \renewcommand{\arraystretch}{1.25}
    
    \caption{Summary of findings in the ablation analysis of distillation's subgroup performance (\S\ref{sec:analysis}).}
    \label{tbl:analysis_summary}
    \vspace{2mm}
    
    \resizebox{0.95\linewidth}{!}{
    \begin{tabular}{@{}lp{2in}@{}}
        \toprule
        \textbf{Section} & \textbf{Finding} \\
        \toprule
        \S\ref{sec:distillation-harms-hard} & Worst-$k$ class error {\color{red}hurt} under self-distillation on Imagenet \\
        \S\ref{sec:ablation-architecture} & Similar results {\color{red}do} hold for settings beyond self-distillation \\ %
        \S\ref{sec:ablation-dataset} & Similar results do {\color{NavyBlue}not} hold for ``easy'' datasets, e.g., CIFAR100 \\
        \bottomrule
    \end{tabular}
    \quad
    \begin{tabular}{@{}lp{2in}@{}}
        \toprule
        \textbf{Section} & \textbf{Finding} \\
        \toprule        
        \S\ref{sec:ablation-imbalance} & Similar results {\color{red}do} hold for long-tailed versions of ``easy'' datasets \\
        \S\ref{sec:why-harm} & Teacher confidently mispredicts on affected subgroups \\
        \S\ref{sec:subgroup-justification} & Worst-$k$ subgroup error {\color{red}hurt} under distillation on fairness datasets \\
        \bottomrule
    \end{tabular}
    }
\end{table}

\subsection{Distillation can help average performance}

To begin, we consider the effect of distillation on a standard image classification benchmark,
namely,
ImageNet.
We employ a self-distillation~\citep{Furlanello:2018} setup,
with 
ResNet-34
teacher and student models,
trained with standard hyperparameter choices (see Appendix \ref{app:experiments}). 
Following~\citet{Cho:2019}, we use early stopping on the teacher model. %

We now ask: what is the impact of distillation on %
the student's \emph{overall} and \emph{per-class} performance?
The first question %
has an expected answer:
distillation improves the student's average accuracy by
$+0.4\%$ (see Table~\ref{tbl:imagenet_architecture_sweep}).
Judged by this conventional metric, distillation is thus a success.

\subsection{Distillation can hurt subgroup performance}
\label{sec:distillation-harms-hard}

\begin{figure}[!t]
\begin{minipage}{\textwidth}
    \begin{minipage}[t]{0.47\textwidth}
    \vspace{0pt}

        \centering
        
        \resizebox{0.7\linewidth}{!}{
            \includegraphics[scale=0.125]{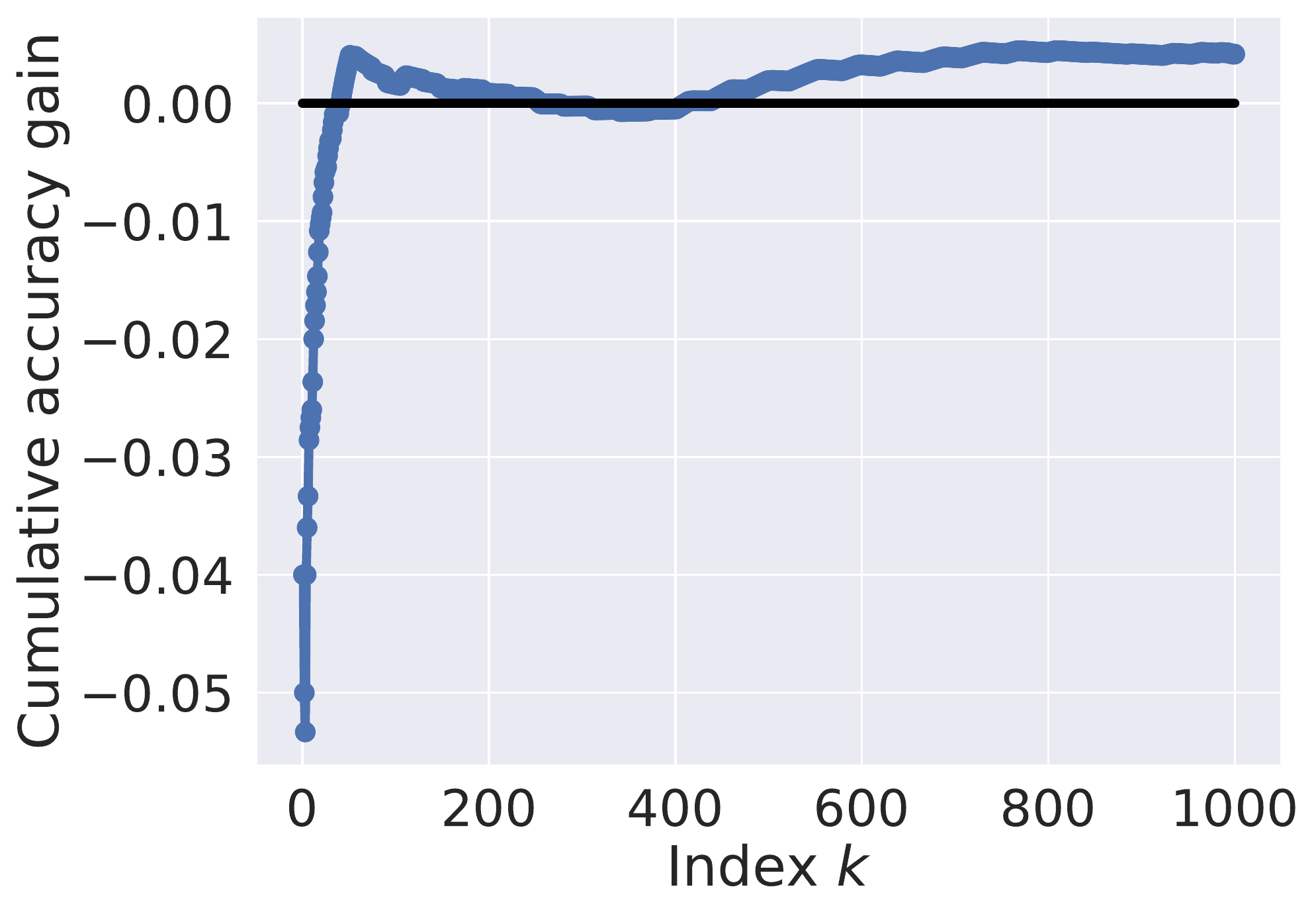}
        }
        \vspace{-2mm}
        \captionof{figure}{
        Cumulative gain of ResNet-34 self-distillation on ImageNet.
        For index $k$, we compute the gain in average accuracy over the $k$ worst classes~\eqref{eqn:top-k}.
        While average accuracy ($k = 1000$) improves by $+0.4\%$,
        for $k \leq 40$, distillation \emph{harms} over the one-hot model.
        }
        \label{fig:imagenet_scatter}
    \end{minipage}
    \qquad
    \begin{minipage}[t]{0.47\textwidth}
    \vspace{0pt}
        
        \resizebox{0.95\linewidth}{!}{
        \begin{tabular}{llll}
            \toprule
            \textbf{Teacher \newline depth} & \textbf{Student \newline  depth} & \textbf{Average \newline accuracy} & \textbf{Worst-$10$ accuracy} \\
            \toprule
            50 & 50 & \helps{+0.39} & \harms{-0.43} \\
            50 & 34 & \helps{+0.39} & \harms{-2.05}  \\
            50 & 18 & \harms{-0.09} & \harms{-1.00} \\
            34 & 34 & \helps{+0.42} & \harms{-2.60} \\
            34 & 18 & \helps{+0.15} & \harms{-3.60} \\
            18 & 18 & \harms{-0.13} & \harms{-3.80} \\
            \bottomrule
        \end{tabular}%
        }
        
        \captionof{table}{Summary of effect of distillation on different teacher and student architectures considered for the ImageNet dataset. Worst-$10$ accuracy is the average accuracy of the worst $10$ classes (cf.~\eqref{eqn:top-k}). 
        The comparison is with respect to the one-hot (i.e., non-distilled) student.
        }
        \label{tbl:imagenet_architecture_sweep}
    \end{minipage}
\end{minipage}
\end{figure}

A more nuanced picture emerges when we break down the source of the above improvement.
\newedit{We compute the per-class accuracies for the one-hot and distillation models, to understand how the overall gains of distillation are distributed.
Figure~\ref{fig:imagenet_scatter} shows that
these gains are non-uniform:
distillation in fact \emph{hurts} the worst-$k$ class performance~\eqref{eqn:top-k} for $k \leq 40$.
(See Appendix~\ref{app:experiments} for a detailed per-class breakdown.)
Quantitatively, 
Table~\ref{tbl:imagenet_architecture_sweep} confirms that
distillation 
{worsens} the worst-10 class accuracy~\eqref{eqn:top-k} 
by 
$-2.6\%$.
Thus, distillation may harm the student on 
classes that it already finds difficult.}

\newedit{Given that average accuracy improves, 
it is worth asking whether the above is a cause for concern:
does it matter that performance on subgroups
corresponding to the ``hardest'' classes
suffers?
While ultimately a domain-specific consideration,
in general exacerbating subgroup errors may lead to issues from the fairness perspective.
Indeed, we shall see that distillation can also harm in settings where the subgroups correspond to sensitive variables;
we discuss this further in~\S\ref{sec:subgroup-justification}.
}

At this stage,
it is apposite to ask whether the above is an isolated finding,
or indicative of a deeper issue.
We thus study each of the following in turn:
\begin{enumerate*}[itemsep=0pt,topsep=0pt,leftmargin=16pt,label=(\roman*)]
    \item does the finding hold in settings beyond self-distillation?
    
    \item does the finding hold for other datasets, or is it simply due to the idiosyncrasies of ImageNet?
    
    \item what are some general characteristics of settings where the problem is manifest?
\end{enumerate*}

\subsection{Is distillation biased by the model size?}
\label{sec:ablation-architecture}

Having begun with a self-distillation setup,
we now demonstrate that similar findings hold when the student and teacher architectures differ.
Continuing with the ImageNet dataset,
in Table~\ref{tbl:imagenet_architecture_sweep}, we report statistics for the overall average accuracy and average accuracy over the worst 10 classes, when varying teacher and student architectures.
The detrimental effect of distillation on hard class performance holds across all scenarios:
thus, our earlier results were not specific to self-distillation.

For self-distillation settings, 
smaller models appear to incur greater losses on the worst-class error.
When distilling between different architectures (e.g., from ResNet-50 to ResNet-18),
we observe
that 
even average accuracy may not improve, as noted in~\citet{Cho:2019}.
There is however no clear trend between
the 
difference in 
architectures
and
drop in worst class performance.

\begin{table*}[!t]
    \renewcommand{\arraystretch}{1.25}
    \centering
    
    \caption{Effect of distillation on different 
    image classification
    datasets. 
    The right column contains long-tailed ({\bf LT}) versions of the datasets on the left.
    We perform self-distillation with a ResNet-56 for the CIFAR datasets. %
    Distillation both helps the average
    while not harming the worst-$1$ class accuracy~\eqref{eqn:top-k}.
    However, 
    long-tailed version of %
    {CIFAR-100} %
    restores the trend of worst-$1$ class accuracy suffering.
    Note that distillation on Imagenet-LT is neutral for worst-$1$ class accuracy due to both distillation and the baseline yielding $0\%$ accuracy for the hardest classes.}
    \label{tbl:dataset_summary}
    
    \vspace{2mm}
    
    \resizebox{0.95\linewidth}{!}{
    \begin{tabular}{lll}
        \toprule
        \textbf{Dataset} & \textbf{Average accuracy} & \textbf{Worst-$1$  accuracy}\\ %
        \toprule
        CIFAR-10     & \helps{+0.55} & \helps{+0.90} \\
        CIFAR-100     & \helps{+1.93} & \helps{+3.33}\\ %
        ImageNet      & \helps{-0.03} & \harms{-1.00}\\ %
        \bottomrule
    \end{tabular}%
    \qquad
    \begin{tabular}{llll}
        \toprule
        \textbf{Dataset} & \textbf{Average accuracy} & \textbf{Worst-$1$  accuracy } & \textbf{Worst-$100$  accuracy} \\
        \toprule    
        CIFAR-10-LT & \helps{+1.92} & \helps{+4.40} & N/A\\    
        CIFAR-100-LT & \helps{+2.17} & \harms{-1.46} & N/A\\ %
        ImageNet-LT & \helps{+0.21} & \neutral{0.00} & \harms{-0.32}\\ %
        \bottomrule
    \end{tabular}
    }
\end{table*}

\subsection{Is distillation biased by a large number of classes?}
\label{sec:ablation-dataset}

Having seen that ImageNet consistently demonstrates a performance degradation on certain classes,
both in self- and conventional-distillation
setups,
we now repeat the same analysis on ``smaller'' image classification benchmarks,
namely,
CIFAR-10, CIFAR-100. %
These have fewer labels than ImageNet. %

We return to the self-distillation setup,
using ResNet-56 models on CIFAR. %
(See Appendix for results with varying architectures.)
On these datasets, Table~\ref{tbl:dataset_summary} shows a (perhaps more expected) result:
distillation boosts \emph{both} the average and {worst-$1$ class} performance.
This indicates that, at a minimum, the behaviour of distillation's performance gains are problem-specific;
on CIFAR, distillation appears a complete win for both the average and {subgroup} accuracy.
One plausible hypothesis is that 
the tension between average and {subgroup} performance only manifests on
problems with many labels,
which, informally, might be considered ``harder''.
However, we now show that 
even for problems with relatively few labels,
one may harm ``hard'' class performance if there is \emph{label imbalance}.

\subsection{Is distillation biased by class imbalance?}
\label{sec:ablation-imbalance}

We now consider a 
\emph{long-tail}
setting, where the training label distribution $\Pr( y )$ is highly non-uniform,
so that most labels have only a few associated samples.
Following %
the  long-tail learning literature~\citep{Cui:2019,Cao:2019,Kang:2020},
we construct ``long-tailed'' ({\bf LT}) versions of the above datasets,
wherein the training set is down-sampled so as to achieve a particular label skew.
For ImageNet, we use the long-tailed version from~\citet{Liu:2019}.
For other datasets,
we %
down-sample labels
to follow $\Pr( y = i ) \propto \frac{1}{\mu^i}$ for constant $\mu$ and $i \in [ L ]$~\citep{Cui:2019}.
The ratio of the most to least frequent class is set to $100$.

From Table~\ref{tbl:dataset_summary},
we note
that on both CIFAR-100-LT and ImageNet-LT, accuracy over the hardest classes drops.
The former is particularly interesting, given that the standard CIFAR-100 shows gains amongst the hardest classes.
This provides evidence that 
for ``harder'' problems
---
e.g., where there are insufficiently many samples from which to model a particular class
---
there may be a tension between the average and {subgroup} performance.

As a qualifying remark, 
CIFAR-10-LT sees the hardest class \emph{improve} upon distillation. %
Thus, label rarity by itself is not predictive of whether distillation harms;
intuitively, classes may be learnable even given a few samples, and thus might see gains under distillation.

\subsection{Why does distillation hurt certain subgroups?}
\label{sec:why-harm}

The above has established that in a range of scenarios,
distillation can hurt 
performance on subgroups defined by individual classes.
However,
a firm understanding of 
\emph{why} this happens remains elusive.
To study this, 
we consider ResNet-56 self-distillation on CIFAR-100-LT
---
which showed a stark gap between the average and subgroup (i.e., {worst-$1$ class}) performance
---
and 
dissect the logits of the teacher
and distilled student. 
(See the Appendix for plots where the teacher and student architectures differ.)
Across classes, we seek to understand:
\begin{enumerate*}[itemsep=0pt,topsep=0pt,leftmargin=16pt,label=(\roman*)]
    \item how aligned are the student and teacher \emph{accuracies}?
    \item how reliable are the models' \emph{probability estimates}?
    \item how do the models' \emph{confidences} behave?
\end{enumerate*}

\begin{figure*}[!t]
    \centering

    \resizebox{\linewidth}{!}{
    \subfigure[Accuracy.]{
    \includegraphics[scale=0.35]{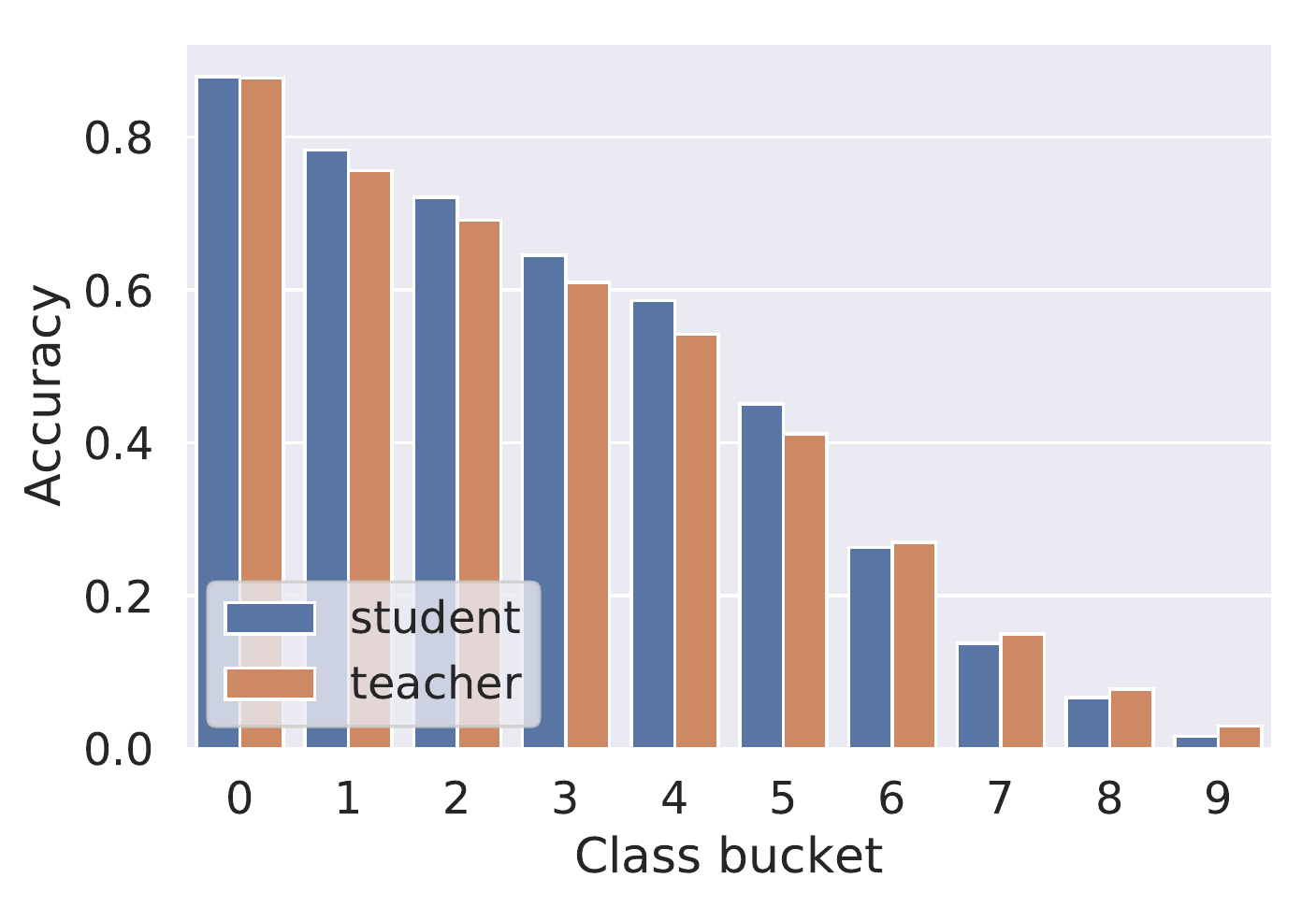}%
    } 
    \hfill
    \subfigure[Log-loss.]{
    \includegraphics[scale=0.35]{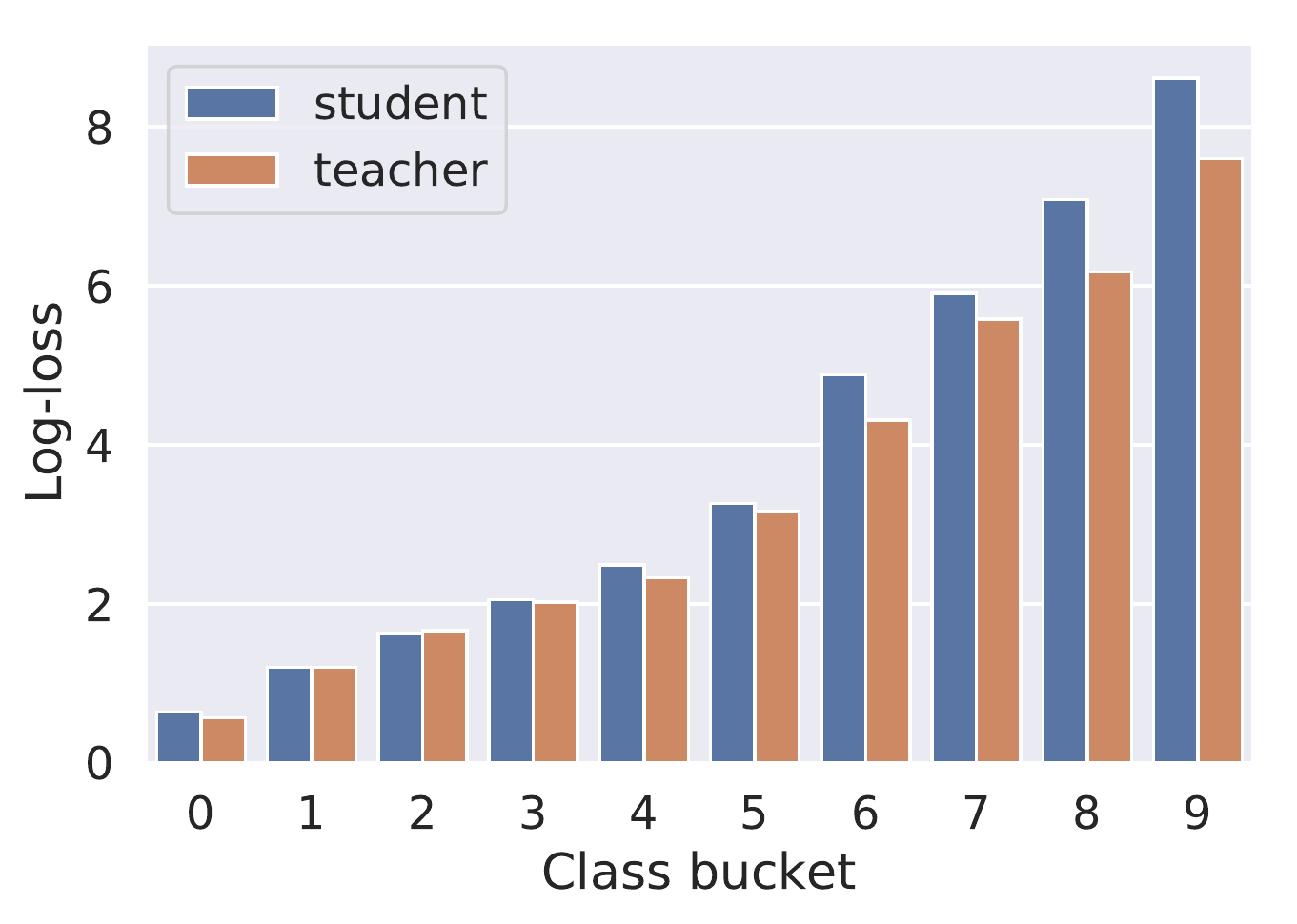}%
    }
    \hfill
    \subfigure[Margin.]{
    \includegraphics[scale=0.35]{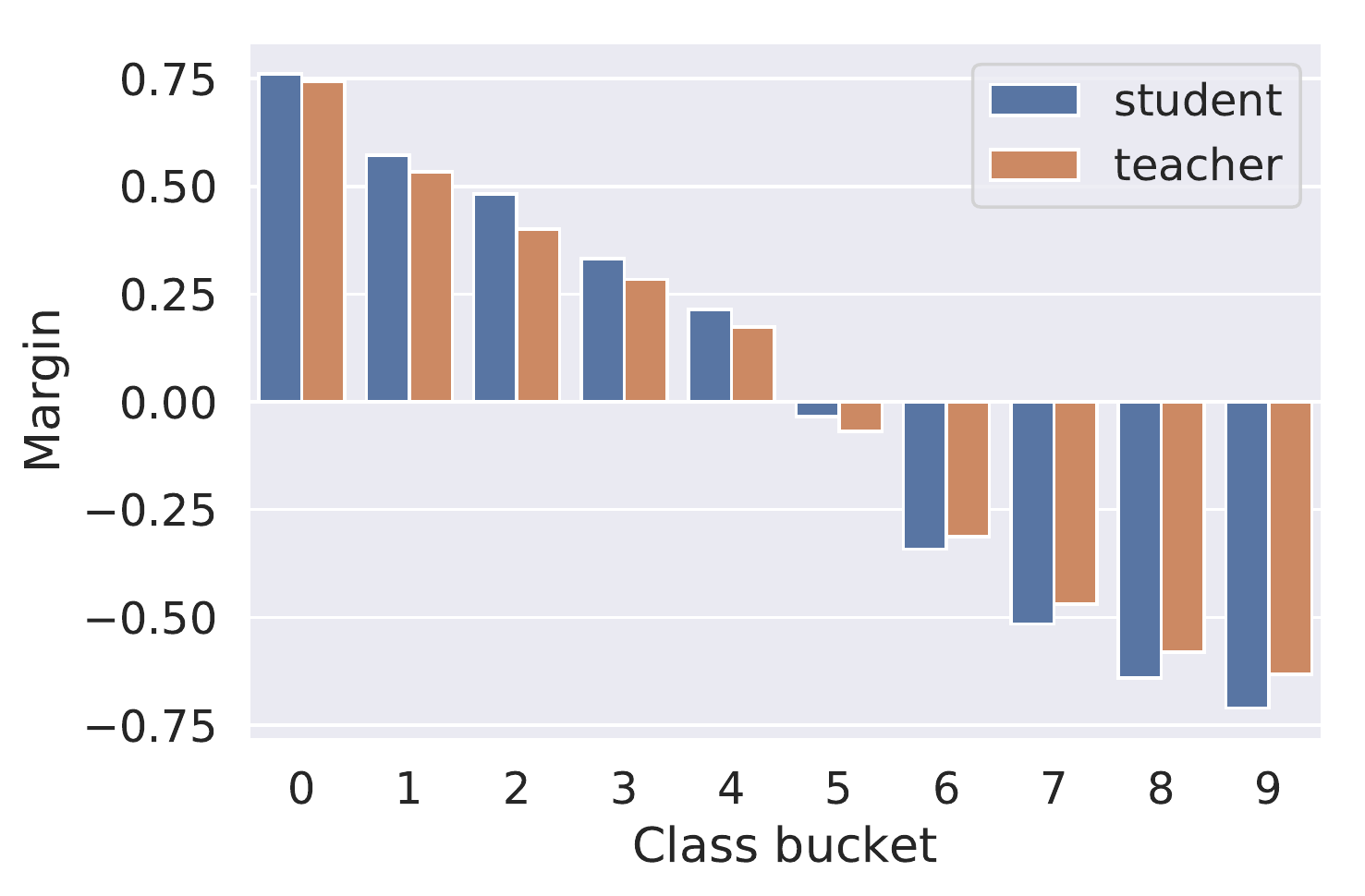}%
    }
    }

    \caption{Logit statistics
    on CIFAR-100 LT,
for the teacher and distilled student under a self-distillation setup (ResNet-56 $\rightarrow$ ResNet-56).
We show the average log-loss, accuracy, and margin of the models on $10$ class buckets:
these are created by sorting the $100$ classes according to the {teacher} accuracy, and then creating $10$ groups of classes.
As expected, the student follows the general trend of the teacher model.
Strikingly, we observe that the teacher model tends to systematically \emph{confidently mispredict} samples in the higher buckets,
thus incurring a \emph{negative} margin;
such misplaced confidence is largely transferred to the student,
whose accuracy suffers on such buckets. 
\newedit{Note that we consider statistics on the \emph{test} set from the teacher and the distilled student, and \emph{not} on the train set.}
    }
    \label{fig:fair_dist_cifar100_lt_logit_breakdown}
\end{figure*}

For a \emph{test}\footnote{\newedit{The choice of test, rather than train, example is crucial: an overparameterised teacher will likely correctly predict \emph{all} training samples, thus rendering the above statistics of limited use. To leverage the insights from the above analysis in practice, we shall use a holdout set that can be carved out from the training set.}} example $(x, y)$ and predicted label distribution $p( x ) \in \Delta_{L}$,
we thus compute 
each models' accuracy,
log-loss $\ell_{\rm log}( y, p( x ) ) = -\log p_y( x )$,
and
\emph{margin}~\citep{Koltchinskii:2002} $\ell_{\rm marg}( y, p( x ) ) = p_y(x) - \max_{y' \neq y} p_y'(x)$.
Note that the latter may be negative if the model predicts the incorrect label for the example.
Figure~\ref{fig:fair_dist_cifar100_lt_logit_breakdown}
shows these metrics on $10$ class buckets:
these are created by sorting the $100$ classes according to the teacher accuracy, and then creating $10$ buckets of classes.
Within each bucket, we compute the average of the metric specified above.

Remarkably, for 5 out of 10 class buckets, average margins are \emph{negative}, suggesting that the teacher is often 
\emph{wrong yet confident} in predicting these classes.
On these buckets, the student accuracy generally worsens compared to the teacher.
Further, log-loss increases across all buckets (including those where accuracy improves), 
indicating reduced confidence \newedit{in the true class} of the distilled student. 
This points at a potential source of the poorer performance on the worst-$1$ accuracy.
\newedit{Recall that the distilled student's aim is to mimic the teacher's logits on the \emph{training} samples.
This is a proxy to the student's true goal, which is mimicking these logits on \emph{test} samples, 
so as to attain similar generalisation performance as the teacher.
When such generalisation happens,
the student can thus be expected to roughly inherit the teacher's per-class performance;
in settings like the above, this unfortunately implies it will perform poorly on those classes with negative teacher margin.}

To verify the above holds more generally, we repeat this analysis
for
other datasets; see Appendix~\ref{app:experiments}.

\subsection{Beyond classes: other choices of subgroups}
\label{sec:subgroup-justification}

\newedit{Our analysis thus far has focused on subgroups defined by classes.
This choice is of natural importance 
for long-tailed problems, 
where it is important to ensure good model performance on rare classes~\citep{Kang:2020}.
In other problems, different choices of subgroups may be appropriate.
For example, in problems arising in fairness,
one may define subgroups based on certain sensitive attributes (e.g., sex, race).
In such settings, does one similarly see varying gains from distillation across subgroups?

We confirm this can indeed hold on the UCI Adult dataset using random forest models (details in Appendix~\ref{app:adult}).
This data involves the task of predicting if an individual's income is $\geq 50$K or not,
and possesses subgroups defined by the individual's race and sex.
Akin to the preceding results, we find that distillation can significantly improve \emph{overall} accuracy,
at the expense of \emph{degrading} accuracy on certain rare subgroups, e.g., Black women; see Table~\ref{tbl:adult_concise}, and Table~\ref{tbl:adult_race_sex} (Appendix).
This further corroborates our basic observation on the non-uniform distribution of distillation's gains.

\begin{table}[!t]
    \renewcommand{\arraystretch}{1.25}
    \centering
    
    \caption{Random forest distillation on UCI Adult dataset helps overall, but hurts on certain subgroups.}
    \label{tbl:adult_concise}
    
    \resizebox{0.5\linewidth}{!}{
    \begin{tabular}{lll}
        \toprule
        \textbf{Average accuracy} & \textbf{Asian Male accuracy} & \textbf{Black Female accuracy} \\ %
        \toprule
        \helps{+3.10} & \harms{-5.94} & \harms{-2.38} \\
        \bottomrule
    \end{tabular}%
    }
\end{table}

A distinct notion of subgroup was recently considered in~\citet{Zhou:2021},
who identified the impact of certain ``regularisation samples'' on distillation.
These are a subset of training samples which were seen to degrade the \emph{overall} performance of distillation.
It is of interest whether such a subgroup relates to
our previously studied subgroups of
``hard'' classes; 
e.g.,
is there an abundance of regularisation samples in such subgroups,
which might explain the poor performance of distillation?
In Appendix~\ref{app:regularisation_samples}, we study the relationship between regularisation samples, 
and the per-label subgroups from our analysis;
we find that, in general, these may be complementary notions.
We further analyze the effect of the technique proposed in~\citet{Zhou:2021} on average and {subgroup} accuracies in~\S\ref{sec:experiments}.}

\section{Mitigating subgroup degradation under distillation}
\label{sec:methods}

We have seen that distillation's gains in average accuracy can be at the expense of degradation in subgroup accuracy.
The preceding analysis indicates that this behaviour is potentially a result of the teacher \emph{confidently mispredicting} on some subgroups.
We now study simple means
of correcting distillation to prevent such undesirable behaviour.
\newedit{In the following, for concreteness and simplicity, we focus on subgroups that are given by the individual classes.}

\subsection{Distillation with adaptive mixing weights}
\label{sec:ada-alpha}
In~\S\ref{sec:why-harm},
we saw that distillation can hurt on classes where the teacher is inherently inaccurate.
Such inaccuracy may in fact be \emph{amplified} by the student, which is hardly desirable.
An intuitive fix is to simply rely less on the teacher for classes where it performs poorly, or is otherwise not confident;
instead, the student can simply fall back onto the one-hot training labels themselves.
Formally, 
for \emph{per-class} mixing weights
$( \alpha_1, \ldots, \alpha_L ) \in [0, 1]^L$,
the student can minimise
\begin{equation}
    \label{eqn:adaptive-mix}
    \begin{aligned}
        \bar{R}_{\mathrm{dist}}( f ) = \frac{1}{N} \sum_{n = 1}^{N} \Big[ &( 1 - \alpha_{y_n} ) \cdot \ell( y_n, f( x_n ) ) + \alpha_{y_n} \cdot \sum_{y' \in [L]} p^{\mathrm{t}}_{y'}( x ) \cdot \ell( y', f( x_n ) ) \Big].
    \end{aligned}
\end{equation}
This objective introduces a mixing weight $\alpha_y$ per-class, 
which allows us to weigh between teacher predictions and one-hot labels for each class independently. By contrast, in the standard distillation setup~\eqref{eqn:distillation} we only have a single weight $\alpha$ that is common for all classes. 

How do we choose values for these weights, $\alpha_{y}$? In the standard distillation objective~\eqref{eqn:distillation},
one only needs to tune a single scalar $\alpha$,
which is amenable to, e.g., cross-validation.
By contrast,~\eqref{eqn:adaptive-mix} involves a single scalar for each label,
which makes any attempt at grid search infeasible.
Following the observations in~\S\ref{sec:why-harm},
we propose the following intuitive setting of $\alpha_y$
given teacher predictions $p^{\rm t}$:
\begin{align*}
    \numberthis
    \label{eqn:ada-alpha}
    \alpha_y = \max\left( 0, \mathbb{E}_{x \mid y}\left[ \gamma_{\rm avg}( y, p^{\rm t}( x ) ) \right] \right) \quad
    \gamma_{\rm avg}( y, p^{\rm t}( x ) ) \defEq p^{\rm t}_y( x ) - \frac{1}{L - 1} \sum_{y' \neq y} p^{\rm t}_{y'}( x ).
\end{align*}
In words,
~\eqref{eqn:ada-alpha}
places greater faith in the teacher model for those classes
which it predicts \emph{correctly} with \emph{confidence},
i.e.,
with large average margin $\gamma_{\rm avg}$.
When this margin is \emph{negative}
---
so that the teacher is \emph{incorrect} on average,
which can occur on classes that are rare in the training set
---
we set $\alpha_y = 0$, and completely ignore the teacher predictions.

The above requires estimating the expectation $\mathbb{E}_{x \mid y}\left[ \cdot \right]$, which requires access to a labelled sample.
This may be done using a holdout set; we shall follow this in our subsequent experiments.

\subsection{Distillation with per-class margins}
\label{sec:ada-margin}
Our second approach for improving distillation on harder classes
is to leverage recent developments
in \emph{long-tail learning},
where the goal is to improve performance on rare classes.
Specifically, 
~\citet{Khan:2018,Cao:2019,Tang:2020,Ren:2020,Menon:2020,Wang:2021}
proposed a variant of the softmax cross-entropy
with \emph{variable margins} $\rho_{yy'}$ between label pairs:
\begin{equation}
    \label{eqn:adaptive-margin}
    \ell( y, f( x ) ) = \log\left[ 1 + \sum_{y' \neq y} \rho_{yy'} \cdot e^{f_{y'}( x ) - f_{y}( x )} \right].
\end{equation}
Intuitively, 
this strongly penalises predicting label $y'$ instead of $y$ when $\rho_{yy'}$ is large.
For training label distribution $\pi$,
\citet{Cao:2019} proposed to set $\rho_{yy'} \propto \exp( \pi_{y}^{-1/4} )$,
so that rare labels receive a higher weight when misclassified.
\citet{Khan:2018,Ren:2020,Menon:2020,Wang:2021} showed gains by instead setting $\rho_{yy'} \propto \frac{\pi_{y'}}{\pi_{y}}$,
so that rare labels are not confused with common ones.

We adapt such techniques to our setting,
with the intuition
that 
we ought to increase the student penalty for misclassifying those ``hard'' classes that the teacher has difficulty modeling.
We thus choose $\rho_{yy'} = \frac{\alpha_{y'}}{\alpha_{y}}$, where $\alpha_y$ is the adaptive per-class mixing weight from the previous section.
This discourages the model from confusing ``hard'' labels $y$ with ``easy'' labels $y'$, when $\alpha_{y'} > \alpha_{y}$.

\subsection{Relation to existing work}
Previous works considered varying distillation supervision across examples with the aim of improving average accuracy.
In particular,
there have been proposals to weight samples 
 based on the ratio~\citep{Tang:2019,Zhou:2021},
and difference~\citep{Zhang:2020d} between student and teacher score.
Similarly,~\citet{Zhou:2020} proposed to only apply distillation on samples the teacher gets correct.

\section{Results for adaptive distillation methods}
\label{sec:experiments}
\newedit{We now present results that 
further corroborate the potential non-uniform gains of distillation,
and the ability to mitigate this with the techniques of the previous section.
We emphasise here that our goal is expressly \emph{not} to improve over the state-of-the-art in distillation techniques;
rather, we wish to verify the key principles identified in the preceding study,
which considers distillation from a novel angle (i.e., in terms of subgroup rather than average performance).}

\textbf{{Experimental setup}}.
We report results on each of the datasets used in~\S\ref{sec:analysis}:
CIFAR-10, CIFAR-100, ImageNet; %
and long-tailed ({\bf LT}) versions of the same.
For brevity, we report results for a self-distillation regime.
(For results with varying architectures, see the Appendix.)
Thus, for each dataset, we train a one-hot teacher ResNet model,
which is distilled to a student ResNet model of the same depth.
We use ResNet-56 models for CIFAR, and ResNet-50 models for all other datasets.
We employ the same hyper-parameters as used in~\S\ref{sec:analysis}, except resorting to non-early stopped teachers for consistency across datasets;
see the Appendix for details.

We compare:
\begin{enumerate*}[itemsep=0pt,topsep=0pt,leftmargin=16pt,label=(\roman*)]
    \item standard one-hot training of the student
    \item standard distillation, i.e., minimising~\eqref{eqn:distillation}
    \item {\bf AdaAlpha}, our proposed distillation 
    objective
    with adaptive mixing between one-hot and teacher labels~\eqref{eqn:adaptive-mix},
    and $\alpha$ as per~\eqref{eqn:ada-alpha}
    \item {\bf AdaMargin}, our proposed distillation 
    objective
    with adaptive margins~\eqref{eqn:adaptive-margin},
    and $\rho_{yy'} = \frac{\alpha_{y'}}{\alpha_{y}}$.
\end{enumerate*}
We summarise the effect of each method 
by reporting the following
per-class accuracy statistics: 
\begin{enumerate*}[itemsep=-4pt,topsep=0pt,leftmargin=16pt,label=(\roman*)]
    \item the standard \emph{mean} accuracy over all classes;
    \item the accuracy over the \emph{worst-$1$} class; and
    \item the mean accuracy over \emph{worst-$10$} (and \emph{worst-$100$} for the LT datasets) classes.
\end{enumerate*}

\newedit{For the Ada-* methods, per~\S\ref{sec:methods}, creating the label-dependent $\alpha_y$ requires estimating the generalisation performance of the teacher.
To do this, we create a random 
holdout split of the training set.
For non-LT datasets, we randomly split into 80\% (new train) -- 20\% (dev).
For LT datasets, for each class we hold out $k$ examples into the dev set ($k = 50$ for Imagenet-LT, $k = 20$ for CIFAR-100-LT and CIFAR-10-LT), or half of examples for a class if the total number of per class examples  is at most $2 k$. 
We train an initial teacher on the new train slice of data,
and estimate its per-class performance on the holdout dev slice.
These are used to estimate $\alpha_y$ as per, e.g.,~\eqref{eqn:ada-alpha}.}

\label{s:results_and_discussion}

Table~\ref{tbl:fair_dist_results} summarises the results for all methods.
We make the following observations.

\begin{table*}[!t]
    \centering
    
    \renewcommand{\arraystretch}{1.25}
    
    \caption{
    Summary of student's average accuracy using one-hot and distilled labels.
    \emph{Worst $k$} denotes accuracy over the worst $k$ classes (for $k \in \{1, 10, 100\}$.
    The proposed AdaAlpha technique improves mean accuracy over vanilla distillation,
    while improving worst-class performance over standard distillation.
    Global temperature and adaptive temperatures $\alpha_y$ selected using a held out dev set. 
    \newedit{For AdaMargin on CIFAR-100 LT and ImageNet LT, we observed divergence during training, presumably due to this method being sensitive to the selection of hyperparameters, which in turn are estimated on very small number of examples per class (e.g.,  just a couple of examples on LT datasets).}
    }
    \label{tbl:fair_dist_results}
    \vspace{2mm}
    
    \resizebox{1.0\linewidth}{!}{
    \begin{tabular}{@{}lllll@{}}
        \toprule
        \multirow{2}{*}{\textbf{Dataset}}  & \multirow{2}{*}{\textbf{Method}} &
        \multicolumn{3}{l}{\textbf{Per-class accuracy statistics}}\\
        & &
        \textbf{Mean} & \textbf{Worst-1} & \textbf{Worst-10}\\
\toprule
\midrule
CIFAR-10 & One-hot & $93.94$ & $85.97$ & $93.94$\\
& Distillation & $\mathbf{94.49}$ & $\mathbf{86.87}$ & $\mathbf{94.49}$\\
& AdaAlpha & $94.29$ & $86.41$ & $94.29$\\
& AdaMargin & $94.32$ & $86.77$ & $94.32$\\
 \midrule
 \midrule
CIFAR-100 & One-hot & $73.31$ & $45.67$ & $52.12$\\
& Distillation & $75.24$ & $49.00$ & $54.65$\\
& AdaAlpha & $\mathbf{75.43}$ & $49.33$ & $56.42$\\
& AdaMargin & $75.15$ & $\mathbf{51.33}$ & $\mathbf{56.62}$\\
 \midrule
 \midrule
ImageNet & One-hot & $76.38$ & $\mathbf{14.00}$ & $23.64$\\
& Distillation & $76.35$ & $13.00$ & $22.02$\\
& AdaAlpha & $\mathbf{76.57}$ & $13.00$ & $23.22$\\
& AdaMargin & $76.36$ & $\mathbf{14.00}$ & $23.20$\\
\midrule
  \bottomrule
\end{tabular}
\qquad
    \begin{tabular}{@{}llllll@{}}    
        \toprule
        \multirow{2}{*}{\textbf{Dataset}}  & \multirow{2}{*}{\textbf{Method}} &
        \multicolumn{4}{l}{\textbf{Per-class accuracy statistics}}\\
        & &
        \textbf{Mean} & \textbf{Worst-1} & \textbf{Worst-10} & \textbf{Worst-100}\\
\toprule

CIFAR-10 LT & One-hot & $75.97$ & $60.33$ & N/A & N/A\\
& Distillation & $77.89$ & $64.73$ & N/A & N/A\\
& AdaAlpha & $78.02$ & $64.70$ & N/A & N/A\\
& AdaMargin & $\mathbf{78.86}$ & $\mathbf{65.53}$ & N/A & N/A\\
\midrule
\midrule
CIFAR-100 LT & One-hot & $43.22$ & $0.00$ & $2.33$ & N/A\\ %
& Distillation & $45.39$ & $0.00$ & $0.87$ & N/A\\ %
& AdaAlpha & $\mathbf{48.57}$ & $\mathbf{0.67}$ & $\mathbf{4.20}$ & N/A\\ %
& AdaMargin\text{*} & - & - & - & -\\ %
\midrule
\midrule
ImageNet LT & One-hot & $45.41$ & $0.00$ & $0.00$ & $\mathbf{1.39}$\\ %
&  Distillation & $45.98$ & $0.00$ & $0.00$ & $\mathbf{1.10}$\\ %
& AdaAlpha & $\mathbf{46.15}$ & $0.00$ & $0.00$ & $1.08$\\ %
& AdaMargin\text{*} & - & - & - & -\\ %
\midrule
\bottomrule
    \end{tabular}
    }
\end{table*}

\begin{table}[!t]
    \centering
    
    \renewcommand{\arraystretch}{1.25}
    
    \caption{Ablations of design choices in the proposed methods: First, remove distillation signal from the bottom 10\% of classes, according to confidence in the true label; Second, 
    randomly shuffle per-class $\alpha$ values; Third, weight distillation weight for examples based on student and teacher confidence in the true label \citet{Zhou:2021}. %
    }
    \label{tbl:simple_ablations}

\resizebox{0.5\linewidth}{!}{
 \begin{tabular}{@{}lllll@{}}
        \toprule
        \multirow{2}{*}{\textbf{Dataset}} & \multirow{2}{*}{\textbf{Method}} &
        \multicolumn{3}{l}{\textbf{Per-class accuracy statistics}}\\
        & &
        \textbf{Mean} & \textbf{Worst-1} & \textbf{Worst-10}\\
\toprule
CIFAR-100 & AdaAlpha & $\mathbf{75.52}\pm0.10$ & $\mathbf{49.33}\pm3.09$ & $\mathbf{56.59}\pm0.44$\\
& remove hardest 10\% & $75.40\pm0.04$ & $48.33\pm1.89$ & $55.79\pm1.19$\\
&  shuffle temperatures & $74.56\pm0.93$ & $46.00\pm1.91$ & $53.10\pm1.22$\\
& \citet{Zhou:2021} & $75.14\pm0.29$ & $45.11\pm1.66$ & $53.89\pm0.66$\\
\midrule
\bottomrule
    \end{tabular}
    }
    \vspace{-\baselineskip}
\end{table}

\textbf{AdaAlpha improves mean accuracy over vanilla distillation}.
The proposed AdaAlpha method consistently and significantly improves 
standard
mean accuracy over vanilla distillation.
Thus, AdaAlpha does not sacrifice the gains offered by distillation on average class performance,
which is desirable.
Other techniques sometimes perform slightly worse than standard distillation on this metric;
however, as we now see, this is compensated by gains on other important dimensions.

\textbf{AdaAlpha improves worst-accuracy over distillation}.
The proposed method consistently improves 
the worst-class accuracy compared to standard distillation:
Thus, the technique largely fulfil their design goal of improving performance on ``hard'' classes,
while not overly sacrificing average-case performance.
In most cases, these improve \emph{both} the average and worst-class accuracy, %
indicating that softening the teacher influence can be broadly beneficial.

\textbf{Comparison of AdaAlpha and AdaMargin}.
In %
the Appendix, we report per class statistics for CIFAR-100 LT. 
AdaMargin flattens both the margin and log-loss distributions, reducing confidence on the poorly classified, tail classes. 
AdaAlpha consistently increases log-loss across classes, and improves margins on few buckets, leading to a positive margin on one bucket where all other methods give negative margins.
Intuitively, AdaMargin tries to more aggressively control the worst-class accuracy;
when this succeeds, there is a large payoff,
but there is also greater risk of overfitting.

\textbf{Additional ablations}.
\newedit{
We confirm that the success of AdaAlpha is not immediately replicated by simpler baselines:
{
\begin{enumerate*}[itemsep=0pt,topsep=0pt,leftmargin=16pt,label=(\roman*)]
    \item \emph{remove hardest 10\%}, which removes the distillation loss component on bottom 10\% labels according to the per class margins found using \ref{eqn:ada-alpha}. 
    It helps analyze whether there is any additional gain beyond simply removing teacher's supervision where it is arguably wrong. %

    \item \emph{shuffle temperatures}, which randomly shuffles the per-class $\alpha_y$ values used in AdaAlpha. 
    This determines whether 
    the precise choice of which labels to up- or down-weight is important;
    
    \item the adaptive distillation scheme of~\citet{Zhou:2021}, where distillation is weighted differently across examples depending on the teacher and student scores.
\end{enumerate*}
}

In Table~\ref{tbl:simple_ablations}, we find that the first two methods work worse than the proposed AdaAlpha method, indicating that the precise choice of which labels to up- or down-weight is important,
and that it does not suffice to merely ignore the teacher on entire subgroups.
The adaptive distillation scheme~\citet{Zhou:2021} is also not as effective as AdaAlpha;
see Appendix for more such results.}

\section{Discussion and other approaches}
\label{sec:conclusion}

Our goal of ensuring equitable performance across classes can be seen as encouraging \emph{fairness}
across subgroups defined by the classes.
This is subtly different to the classical fairness literature~\citep{Calders:2010,Dwork:2012,Hardt:2016}, 
wherein the subgroups are defined by certain \emph{sensitive attributes}.
Broadly, fairness techniques
attempt to learn models that predict the target label \emph{accurately},
but the subgroup label \emph{poorly};
these are inadmissible for our setting, wherein the two labels exactly coincide.
Ensuring fairness across subgroups defined by the classes
has been studied in~\citet{Mohri:2019,Williamson:2019,Sagawa:2020}, 
who proposed algorithms to explicitly minimise the \emph{worst-class} (as opposed to the average) loss.
Adapting such algorithms to the distillation setting is of interest for future work.
\newedit{More broadly, the intent of the analysis in this paper is to better understand settings where distillation can implicitly hurt certain under-represented subgroups.
For societal applications, it is also important to verify that in settings with fairness \emph{constraints},
the proposed techniques are effective in mitigating such degradation, and do not introduce unforeseen implicit biases.}

\bibliography{main}

\begin{thebibliography}{46}
\providecommand{\natexlab}[1]{#1}
\providecommand{\url}[1]{\texttt{#1}}
\expandafter\ifx\csname urlstyle\endcsname\relax
  \providecommand{\doi}[1]{doi: #1}\else
  \providecommand{\doi}{doi: \begingroup \urlstyle{rm}\Url}\fi

\bibitem[Allen{-}Zhu and Li(2020)]{AllenZhu:2020}
Zeyuan Allen{-}Zhu and Yuanzhi Li.
\newblock Towards understanding ensemble, knowledge distillation and
  self-distillation in deep learning.
\newblock \emph{CoRR}, abs/2012.09816, 2020.
\newblock URL \url{https://arxiv.org/abs/2012.09816}.

\bibitem[Anil et~al.(2018)Anil, Pereyra, Passos, Ormandi, Dahl, and
  Hinton]{Anil:2018}
Rohan Anil, Gabriel Pereyra, Alexandre Passos, Robert Ormandi, George~E. Dahl,
  and Geoffrey~E. Hinton.
\newblock Large scale distributed neural network training through online
  distillation.
\newblock In \emph{International Conference on Learning Representations}, 2018.

\bibitem[Bucil\v{a} et~al.(2006)Bucil\v{a}, Caruana, and
  Niculescu-Mizil]{Bucilua:2006}
Cristian Bucil\v{a}, Rich Caruana, and Alexandru Niculescu-Mizil.
\newblock Model compression.
\newblock In \emph{Proceedings of the 12th ACM SIGKDD International Conference
  on Knowledge Discovery and Data Mining}, KDD '06, pages 535--541, New York,
  NY, USA, 2006. ACM.

\bibitem[Buda et~al.(2017)Buda, Maki, and Mazurowski]{Buda:2017}
Mateusz Buda, Atsuto Maki, and Maciej~A. Mazurowski.
\newblock A systematic study of the class imbalance problem in convolutional
  neural networks.
\newblock \emph{arXiv:1710.05381 [cs, stat]}, October 2017.

\bibitem[Buolamwini and Gebru(2018)]{Buolamwini:2018}
Joy Buolamwini and Timnit Gebru.
\newblock Gender shades: Intersectional accuracy disparities in commercial
  gender classification.
\newblock In Sorelle~A. Friedler and Christo Wilson, editors, \emph{Conference
  on Fairness, Accountability, and Transparency}, volume~81 of
  \emph{Proceedings of Machine Learning Research}, pages 77--91, New York, NY,
  USA, 23--24 Feb 2018. PMLR.

\bibitem[Calders and Verwer(2010)]{Calders:2010}
Toon Calders and Sicco Verwer.
\newblock Three {N}aive {B}ayes approaches for discrimination-free
  classification.
\newblock \emph{Data Mining and Knowledge Discovery}, 21\penalty0 (2):\penalty0
  277--292, 2010.

\bibitem[Cao et~al.(2019)Cao, Wei, Gaidon, Ar{\'{e}}chiga, and Ma]{Cao:2019}
Kaidi Cao, Colin Wei, Adrien Gaidon, Nikos Ar{\'{e}}chiga, and Tengyu Ma.
\newblock Learning imbalanced datasets with label-distribution-aware margin
  loss.
\newblock In \emph{Advances in Neural Information Processing Systems 32}, pages
  1565--1576, 2019.

\bibitem[{Cho} and {Hariharan}(2019)]{Cho:2019}
J.~H. {Cho} and B.~{Hariharan}.
\newblock On the efficacy of knowledge distillation.
\newblock In \emph{2019 IEEE/CVF International Conference on Computer Vision
  (ICCV)}, pages 4793--4801, 2019.

\bibitem[Chzhen et~al.(2019)Chzhen, Denis, Hebiri, Oneto, and
  Pontil]{Chzhen:2019}
Evgenii Chzhen, Christophe Denis, Mohamed Hebiri, Luca Oneto, and Massimiliano
  Pontil.
\newblock Leveraging labeled and unlabeled data for consistent fair binary
  classification.
\newblock In H.~Wallach, H.~Larochelle, A.~Beygelzimer, F.~d'Alch\'{e} Buc,
  E.~Fox, and R.~Garnett, editors, \emph{Advances in Neural Information
  Processing Systems 32}, pages 12760--12770. Curran Associates, Inc., 2019.

\bibitem[Cui et~al.(2019)Cui, Jia, Lin, Song, and Belongie]{Cui:2019}
Yin Cui, Menglin Jia, Tsung-Yi Lin, Yang Song, and Serge Belongie.
\newblock Class-balanced loss based on effective number of samples.
\newblock In \emph{CVPR}, 2019.

\bibitem[Dao et~al.(2021)Dao, Kamath, Syrgkanis, and Mackey]{Dao:2021}
Tri Dao, Govinda~M Kamath, Vasilis Syrgkanis, and Lester Mackey.
\newblock Knowledge distillation as semiparametric inference.
\newblock In \emph{International Conference on Learning Representations}, 2021.
\newblock URL \url{https://openreview.net/forum?id=m4UCf24r0Y}.

\bibitem[Dwork et~al.(2012)Dwork, Hardt, Pitassi, Reingold, and
  Zemel]{Dwork:2012}
Cynthia Dwork, Moritz Hardt, Toniann Pitassi, Omer Reingold, and Richard Zemel.
\newblock Fairness through awareness.
\newblock In \emph{Innovations in Theoretical Computer Science Conference
  (ITCS)}, pages 214--226, 2012.

\bibitem[Furlanello et~al.(2018)Furlanello, Lipton, Tschannen, Itti, and
  Anandkumar]{Furlanello:2018}
Tommaso Furlanello, Zachary~Chase Lipton, Michael Tschannen, Laurent Itti, and
  Anima Anandkumar.
\newblock Born-again neural networks.
\newblock In \emph{Proceedings of the 35th International Conference on Machine
  Learning, {ICML} 2018, Stockholmsm{\"{a}}ssan, Stockholm, Sweden, July 10-15,
  2018}, pages 1602--1611, 2018.

\bibitem[Guo et~al.(2017)Guo, Pleiss, Sun, and Weinberger]{Guo:2017}
Chuan Guo, Geoff Pleiss, Yu~Sun, and Kilian~Q. Weinberger.
\newblock On calibration of modern neural networks.
\newblock In \emph{Proceedings of the 34th International Conference on Machine
  Learning}, pages 1321--1330, 2017.

\bibitem[Hardt et~al.(2016)Hardt, Price, and Srebro]{Hardt:2016}
Moritz Hardt, Eric Price, and Nathan Srebro.
\newblock Equality of opportunity in supervised learning.
\newblock In \emph{Advances in Neural Information Processing Systems (NIPS)},
  December 2016.

\bibitem[He et~al.(2016)He, Zhang, Ren, and Sun]{he2016deep}
Kaiming He, Xiangyu Zhang, Shaoqing Ren, and Jian Sun.
\newblock Deep residual learning for image recognition.
\newblock In \emph{Proceedings of the IEEE conference on computer vision and
  pattern recognition}, pages 770--778, 2016.

\bibitem[Hinton et~al.(2015)Hinton, Vinyals, and Dean]{Hinton:2015}
Geoffrey~E. Hinton, Oriol Vinyals, and Jeffrey Dean.
\newblock Distilling the knowledge in a neural network.
\newblock \emph{CoRR}, abs/1503.02531, 2015.

\bibitem[Ji and Zhu(2020)]{Ji:2020}
Guangda Ji and Zhanxing Zhu.
\newblock Knowledge distillation in wide neural networks: Risk bound, data
  efficiency and imperfect teacher.
\newblock In Hugo Larochelle, Marc'Aurelio Ranzato, Raia Hadsell,
  Maria{-}Florina Balcan, and Hsuan{-}Tien Lin, editors, \emph{Advances in
  Neural Information Processing Systems}, 2020.

\bibitem[Jones et~al.(2021)Jones, Sagawa, Koh, Kumar, and Liang]{Jones:2021}
Erik Jones, Shiori Sagawa, Pang~Wei Koh, Ananya Kumar, and Percy Liang.
\newblock Selective classification can magnify disparities across groups.
\newblock In \emph{International Conference on Learning Representations}, 2021.

\bibitem[Kang et~al.(2020)Kang, Xie, Rohrbach, Yan, Gordo, Feng, and
  Kalantidis]{Kang:2020}
Bingyi Kang, Saining Xie, Marcus Rohrbach, Zhicheng Yan, Albert Gordo, Jiashi
  Feng, and Yannis Kalantidis.
\newblock Decoupling representation and classifier for long-tailed recognition.
\newblock In \emph{Eighth International Conference on Learning Representations
  (ICLR)}, 2020.

\bibitem[Khan et~al.(2018)Khan, Hayat, Bennamoun, Sohel, and
  Togneri]{Khan:2018}
Salman~H. Khan, Munawar Hayat, Mohammed Bennamoun, Ferdous~A. Sohel, and
  Roberto Togneri.
\newblock Cost-sensitive learning of deep feature representations from
  imbalanced data.
\newblock \emph{IEEE Transactions on Neural Networks and Learning Systems},
  29\penalty0 (8):\penalty0 3573--3587, 2018.
\newblock \doi{10.1109/TNNLS.2017.2732482}.

\bibitem[Koltchinskii and Panchenko(2002)]{Koltchinskii:2002}
V.~Koltchinskii and D.~Panchenko.
\newblock Empirical margin distributions and bounding the generalization error
  of combined classifiers.
\newblock \emph{Ann. Statist.}, 30\penalty0 (1):\penalty0 1--50, 02 2002.

\bibitem[Liu et~al.(2019)Liu, Miao, Zhan, Wang, Gong, and Yu]{Liu:2019}
Ziwei Liu, Zhongqi Miao, Xiaohang Zhan, Jiayun Wang, Boqing Gong, and Stella~X.
  Yu.
\newblock Large-scale long-tailed recognition in an open world.
\newblock In \emph{{IEEE} Conference on Computer Vision and Pattern
  Recognition, {CVPR} 2019, Long Beach, CA, USA, June 16-20, 2019}, pages
  2537--2546. Computer Vision Foundation / {IEEE}, 2019.

\bibitem[Lopez-Paz et~al.(2016)Lopez-Paz, Sch{\"o}lkopf, Bottou, and
  Vapnik]{Lopez-Paz:2016}
D.~Lopez-Paz, B.~Sch{\"o}lkopf, L.~Bottou, and V.~Vapnik.
\newblock Unifying distillation and privileged information.
\newblock In \emph{International Conference on Learning Representations
  (ICLR)}, November 2016.

\bibitem[Loshchilov and Hutter(2017)]{loshchilov2017sgdr}
Ilya Loshchilov and Frank Hutter.
\newblock Sgdr: Stochastic gradient descent with warm restarts, 2017.

\bibitem[Menon et~al.(2013)Menon, Narasimhan, Agarwal, and Chawla]{Menon:2013}
Aditya~Krishna Menon, Harikrishna Narasimhan, Shivani Agarwal, and Sanjay
  Chawla.
\newblock On the statistical consistency of algorithms for binary
  classification under class imbalance.
\newblock In \emph{Proceedings of the 30th International Conference on Machine
  Learning, {ICML} 2013, Atlanta, GA, USA, 16-21 June 2013}, pages 603--611,
  2013.

\bibitem[Menon et~al.(2020)Menon, Jayasumana, Rawat, Jain, Veit, and
  Kumar]{Menon:2020}
Aditya~Krishna Menon, Sadeep Jayasumana, Ankit~Singh Rawat, Himanshu Jain,
  Andreas Veit, and Sanjiv Kumar.
\newblock Long-tail learning via logit adjustment, 2020.

\bibitem[Mobahi et~al.(2020)Mobahi, Farajtabar, and Bartlett]{Mobahi:2020}
Hossein Mobahi, Mehrdad Farajtabar, and Peter~L. Bartlett.
\newblock Self-distillation amplifies regularization in hilbert space, 2020.

\bibitem[Mohri et~al.(2019)Mohri, Sivek, and Suresh]{Mohri:2019}
Mehryar Mohri, Gary Sivek, and Ananda~Theertha Suresh.
\newblock Agnostic federated learning.
\newblock In \emph{International Conference on Machine Learning}, 2019.

\bibitem[Radosavovic et~al.(2018)Radosavovic, Doll{\'{a}}r, Girshick, Gkioxari,
  and He]{Radosavovic:2018}
Ilija Radosavovic, Piotr Doll{\'{a}}r, Ross~B. Girshick, Georgia Gkioxari, and
  Kaiming He.
\newblock Data distillation: Towards omni-supervised learning.
\newblock In \emph{2018 {IEEE} Conference on Computer Vision and Pattern
  Recognition, {CVPR} 2018, Salt Lake City, UT, USA, June 18-22, 2018}, pages
  4119--4128, 2018.

\bibitem[Ren et~al.(2020)Ren, Yu, sheng, Ma, Zhao, Yi, and Li]{Ren:2020}
Jiawei Ren, Cunjun Yu, shunan sheng, Xiao Ma, Haiyu Zhao, Shuai Yi, and
  hongsheng Li.
\newblock Balanced meta-softmax for long-tailed visual recognition.
\newblock In H.~Larochelle, M.~Ranzato, R.~Hadsell, M.~F. Balcan, and H.~Lin,
  editors, \emph{Advances in Neural Information Processing Systems}, volume~33,
  pages 4175--4186. Curran Associates, Inc., 2020.

\bibitem[Sagawa et~al.(2020{\natexlab{a}})Sagawa, Koh, Hashimoto, and
  Liang]{Sagawa:2020}
S.~Sagawa, P.~W. Koh, T.~B. Hashimoto, and P.~Liang.
\newblock Distributionally robust neural networks for group shifts: On the
  importance of regularization for worst-case generalization.
\newblock In \emph{International Conference on Learning Representations
  (ICLR)}, 2020{\natexlab{a}}.

\bibitem[Sagawa et~al.(2020{\natexlab{b}})Sagawa, Raghunathan, Koh, and
  Liang]{Sagawa:2020b}
S.~Sagawa, A.~Raghunathan, P.~W. Koh, and P.~Liang.
\newblock An investigation of why overparameterization exacerbates spurious
  correlations.
\newblock In \emph{International Conference on Machine Learning (ICML)},
  2020{\natexlab{b}}.

\bibitem[Samadi et~al.(2018)Samadi, Tantipongpipat, Morgenstern, Singh, and
  Vempala]{Samadi:2018}
Samira Samadi, Uthaipon Tantipongpipat, Jamie~H Morgenstern, Mohit Singh, and
  Santosh Vempala.
\newblock The price of fair pca: One extra dimension.
\newblock In S.~Bengio, H.~Wallach, H.~Larochelle, K.~Grauman, N.~Cesa-Bianchi,
  and R.~Garnett, editors, \emph{Advances in Neural Information Processing
  Systems}, volume~31, pages 10976--10987. Curran Associates, Inc., 2018.

\bibitem[Sohoni et~al.(2020)Sohoni, Dunnmon, Angus, Gu, and
  R\'{e}]{Sohoni:2020}
N.~Sohoni, J.~Dunnmon, G.~Angus, A.~Gu, and C.~R\'{e}.
\newblock No subclass left behind: Fine-grained robustness in coarse-grained
  classification problems.
\newblock In \emph{To appear in Conference on Neural Information Processing
  Systems (NeurIPS)}, 2020.

\bibitem[Tang et~al.(2020)Tang, Shivanna, Zhao, Lin, Singh, Chi, and
  Jain]{Tang:2020}
Jiaxi Tang, Rakesh Shivanna, Zhe Zhao, Dong Lin, Anima Singh, Ed~H. Chi, and
  Sagar Jain.
\newblock Understanding and improving knowledge distillation.
\newblock \emph{CoRR}, abs/2002.03532, 2020.

\bibitem[Tang et~al.(2019)Tang, Feng, Shao, Kuang, Zhang, and Lu]{Tang:2019}
Shitao Tang, Litong Feng, Wenqi Shao, Zhanghui Kuang, Wayne Zhang, and Zheng
  Lu.
\newblock Learning efficient detector with semi-supervised adaptive
  distillation.
\newblock In \emph{30th British Machine Vision Conference 2019, {BMVC} 2019,
  Cardiff, UK, September 9-12, 2019}, page 215. {BMVA} Press, 2019.
\newblock URL
  \url{https://bmvc2019.org/wp-content/uploads/papers/0145-paper.pdf}.

\bibitem[Van~Horn and Perona(2017)]{VanHorn:2017}
Grant Van~Horn and Pietro Perona.
\newblock The devil is in the tails: Fine-grained classification in the wild.
\newblock \emph{arXiv preprint arXiv:1709.01450}, 2017.

\bibitem[{Van Horn} et~al.(2018){Van Horn}, {Mac Aodha}, Song, Cui, Sun,
  Shepard, Adam, Perona, and Belongie]{iNaturalist}
Grant {Van Horn}, Oisin {Mac Aodha}, Yang Song, Yin Cui, Chen Sun, Alex
  Shepard, Hartwig Adam, Pietro Perona, and Serge Belongie.
\newblock The inaturalist species classification and detection dataset.
\newblock In \emph{2018 IEEE/CVF Conference on Computer Vision and Pattern
  Recognition}, pages 8769--8778, United States, December 2018. Institute of
  Electrical and Electronics Engineers (IEEE).
\newblock ISBN 978-1-5386-6421-6.
\newblock \doi{10.1109/CVPR.2018.00914}.
\newblock URL \url{http://cvpr2018.thecvf.com/}.
\newblock 2018 IEEE/CVF Conference on Computer Vision and Pattern Recognition,
  CVPR 2018 ; Conference date: 18-06-2018 Through 22-06-2018.

\bibitem[Wang et~al.(2021)Wang, Zhang, Zang, Cao, Pang, Gong, Chen, Liu, Loy,
  and Lin]{Wang:2021}
Jiaqi Wang, Wenwei Zhang, Yuhang Zang, Yuhang Cao, Jiangmiao Pang, Tao Gong,
  Kai Chen, Ziwei Liu, Chen~Change Loy, and Dahua Lin.
\newblock Seesaw loss for long-tailed instance segmentation.
\newblock In \emph{Proceedings of the {IEEE} Conference on Computer Vision and
  Pattern Recognition}, 2021.

\bibitem[Williamson and Menon(2019)]{Williamson:2019}
Robert~C. Williamson and Aditya~Krishna Menon.
\newblock Fairness risk measures.
\newblock In \emph{Proceedings of the 36th International Conference on Machine
  Learning, {ICML} 2019, 9-15 June 2019, Long Beach, California, {USA}}, pages
  6786--6797, 2019.

\bibitem[Zhang et~al.(2020)Zhang, Lan, Dai, Zeng, Bai, Chang, and
  Wei]{Zhang:2020d}
Youcai Zhang, Zhonghao Lan, Yuchen Dai, Fangao Zeng, Yan Bai, Jie Chang, and
  Yichen Wei.
\newblock Prime-aware adaptive distillation.
\newblock In Andrea Vedaldi, Horst Bischof, Thomas Brox, and Jan-Michael Frahm,
  editors, \emph{Computer Vision -- ECCV 2020}, pages 658--674. Springer
  International Publishing, 2020.
\newblock ISBN 978-3-030-58529-7.

\bibitem[Zhang and Sabuncu(2020)]{Zhang:2020b}
Zhilu Zhang and Mert~R. Sabuncu.
\newblock Self-distillation as instance-specific label smoothing.
\newblock In Hugo Larochelle, Marc'Aurelio Ranzato, Raia Hadsell,
  Maria{-}Florina Balcan, and Hsuan{-}Tien Lin, editors, \emph{Advances in
  Neural Information Processing Systems 33: Annual Conference on Neural
  Information Processing Systems 2020, NeurIPS 2020, December 6-12, 2020,
  virtual}, 2020.

\bibitem[Zhao et~al.(2020)Zhao, Xiao, Gan, Zhang, and Xia]{Zhao:2020}
Bowen Zhao, Xi~Xiao, Guojun Gan, Bin Zhang, and Shu{-}Tao Xia.
\newblock Maintaining discrimination and fairness in class incremental
  learning.
\newblock In \emph{2020 {IEEE/CVF} Conference on Computer Vision and Pattern
  Recognition}, pages 13205--13214. {IEEE}, 2020.

\bibitem[Zhou et~al.(2021)Zhou, Song, Chen, Zhou, Wang, Yuan, and
  Zhang]{Zhou:2021}
Helong Zhou, Liangchen Song, Jiajie Chen, Ye~Zhou, Guoli Wang, Junsong Yuan,
  and Qian Zhang.
\newblock Rethinking soft labels for knowledge distillation: A
  bias{\textendash}variance tradeoff perspective.
\newblock In \emph{International Conference on Learning Representations}, 2021.

\bibitem[Zhou et~al.(2020)Zhou, Zhuge, Guan, and Liu]{Zhou:2020}
Zaida Zhou, Chaoran Zhuge, Xinwei Guan, and Wen Liu.
\newblock Channel distillation: Channel-wise attention for knowledge
  distillation.
\newblock \emph{CoRR}, abs/2006.01683, 2020.
\newblock URL \url{https://arxiv.org/abs/2006.01683}.

\end{thebibliography}
\bibliographystyle{plainnat}

\appendix
\section*{Checklist}

\begin{enumerate}

\item For all authors...
\begin{enumerate}
  \item Do the main claims made in the abstract and introduction accurately reflect the paper's contributions and scope?
    \answerYes{}
  \item Did you describe the limitations of your work?
    \answerYes{} {\bf\color{gray}Please see conclusion.}
  \item Did you discuss any potential negative societal impacts of your work?
    \answerYes{} {\bf\color{gray}Please see conclusion.}
  \item Have you read the ethics review guidelines and ensured that your paper conforms to them?
    \answerYes{}
\end{enumerate}

\item If you are including theoretical results...
\begin{enumerate}
  \item Did you state the full set of assumptions of all theoretical results?
    \answerNA{}{}
	\item Did you include complete proofs of all theoretical results?
    \answerNA{}
\end{enumerate}

\item If you ran experiments...
\begin{enumerate}
  \item Did you include the code, data, and instructions needed to reproduce the main experimental results (either in the supplemental material or as a URL)?
    \answerNo{}
  \item Did you specify all the training details (e.g., data splits, hyperparameters, how they were chosen)?
    \answerYes{} {\bf\color{gray}Please see Appendix~\ref{app:experiment-details}.}
	\item Did you report error bars (e.g., with respect to the random seed after running experiments multiple times)?
    \answerYes{}
	\item Did you include the total amount of compute and the type of resources used (e.g., type of GPUs, internal cluster, or cloud provider)?
    \answerNo{} {\bf\color{gray}We employ standard benchmark tasks and architectures, and our focus is not on computational aspects. Our new techniques do not increase the complexity of training or inference.}
\end{enumerate}

\item If you are using existing assets (e.g., code, data, models) or curating/releasing new assets...
\begin{enumerate}
  \item If your work uses existing assets, did you cite the creators?
    \answerYes{}
  \item Did you mention the license of the assets?
    \answerNo{}
  \item Did you include any new assets either in the supplemental material or as a URL?
    \answerNo{}
  \item Did you discuss whether and how consent was obtained from people whose data you're using/curating?
    \answerNA{}
  \item Did you discuss whether the data you are using/curating contains personally identifiable information or offensive content?
    \answerNA{}
\end{enumerate}

\item If you used crowdsourcing or conducted research with human subjects...
\begin{enumerate}
  \item Did you include the full text of instructions given to participants and screenshots, if applicable?
    \answerNA{}
  \item Did you describe any potential participant risks, with links to Institutional Review Board (IRB) approvals, if applicable?
    \answerNA{}
  \item Did you include the estimated hourly wage paid to participants and the total amount spent on participant compensation?
    \answerNA{}
\end{enumerate}

\end{enumerate}

\onecolumn

\section{Details of experiments}
\label{app:experiment-details}
\subsection{Architecture}

We use ResNet with batch norm~\citep{he2016deep} for all our experiments with the following configurations. 
For CIFAR, we experiment with ResNet-56 and ResNet-32.
For ImageNet, we use ResNet-50.
We list the architecture configurations in terms of ($\text{n}_\text{layer}$, $\text{n}_\text{filter}$, stride) corresponding to each ResNet block in Table~\ref{tbl:architecture}.

\begin{table*}[!ht]
    \centering
    
    \renewcommand{\arraystretch}{1.25}
    
    \begin{tabular}{ll}
        \toprule
        \textbf{Architecture} &\textbf{Configuration:  [($\text{n}_\text{layer}$, $\text{n}_\text{filter}$, stride)]} \\
        \toprule
        CIFAR ResNet-32 & [(5, 16, 1), (5, 32, 2), (5, 64, 2)] \\ 
        CIFAR ResNet-56 & [(9, 16, 1), (9, 32, 2), (9, 64, 2)] \\
        \hline 
        ImageNet ResNet-18 & [(2, 64, 1), (2, 128, 2), (2, 256, 2), (2, 512, 2)] \\ 
        ImageNet ResNet-34 & [(3, 64, 1), (4, 128, 2), (6, 256, 2), (3, 512, 2)] \\ 
        ImageNet ResNet-50 & [(3, 64, 1), (4, 128, 2), (6, 256, 2), (3, 512, 2)]* \\ 
        \bottomrule
    \end{tabular}
    \caption{ResNet Architecture configurations used in our experiments~\citep{he2016deep}. [*] Note that ImageNet ResNet-50 uses larger blocks with 3 convolutional layers per residual block compared to ResNet-18 and 34. We refer to \citet{he2016deep} for more details.}
    \label{tbl:architecture}
\end{table*}

\subsection{Training set}

For all datasets, we train using SGD and weight decay $10^{-4}$ for CIFAR, and $0.5 \times 10^{-4}$ for Imagenet datasets.
We have the following dataset specific settings.

\textbf{CIFAR-10 and CIFAR-100}.
We train for 450 epochs with 
an initial learning rate of 1.0,
with a linear warmup in the first 15 epochs,
and 
an annealed learning rate schedule. 
We drop the learning rate by a
factor of 10 at epochs number: 200, 300 and 400.
We use a mini-batch size of 1024.
We use SGD with Nesterov momentum of 0.9.

For our distillation experiments we train only with the cross-entropy objective against the teacher's logits.
For each method we find the best temperature from the list of values: $\{1, 2, 3, 4, 5\}$.

\textbf{ImageNet}.
We train for 90 epochs 
with  an initial learning rate of 0.8,
with a linear warmup in the first 5 epochs,
and
an annealed learning rate schedule. 
We drop the learning rate by a
factor of 10 at epochs number: 30, 60 and 80. 
We use a mini-batch size of 1024.

For our distillation experiments we train with the distillation objective as defined in Equation~\ref{eqn:distillation} setting $\alpha=0.2$.
For each method we fix the temperature to $0.9$.

\textbf{Long-tail (LT) datasets}. We follow setup as in the non-long tail version, except for the learning rate schedule, which we change to follow the cosine schedule~\citep{loshchilov2017sgdr}.

\section{Additional Experiments}
\label{app:experiments}

We present additional experiments to those in the body.

\subsection{Further varying datasets and model architectures}

\begin{figure*}[!t]
    \centering

    \resizebox{\linewidth}{!}{
    \subfigure[Accuracy (tail classes).]{
    \includegraphics[scale=0.35]{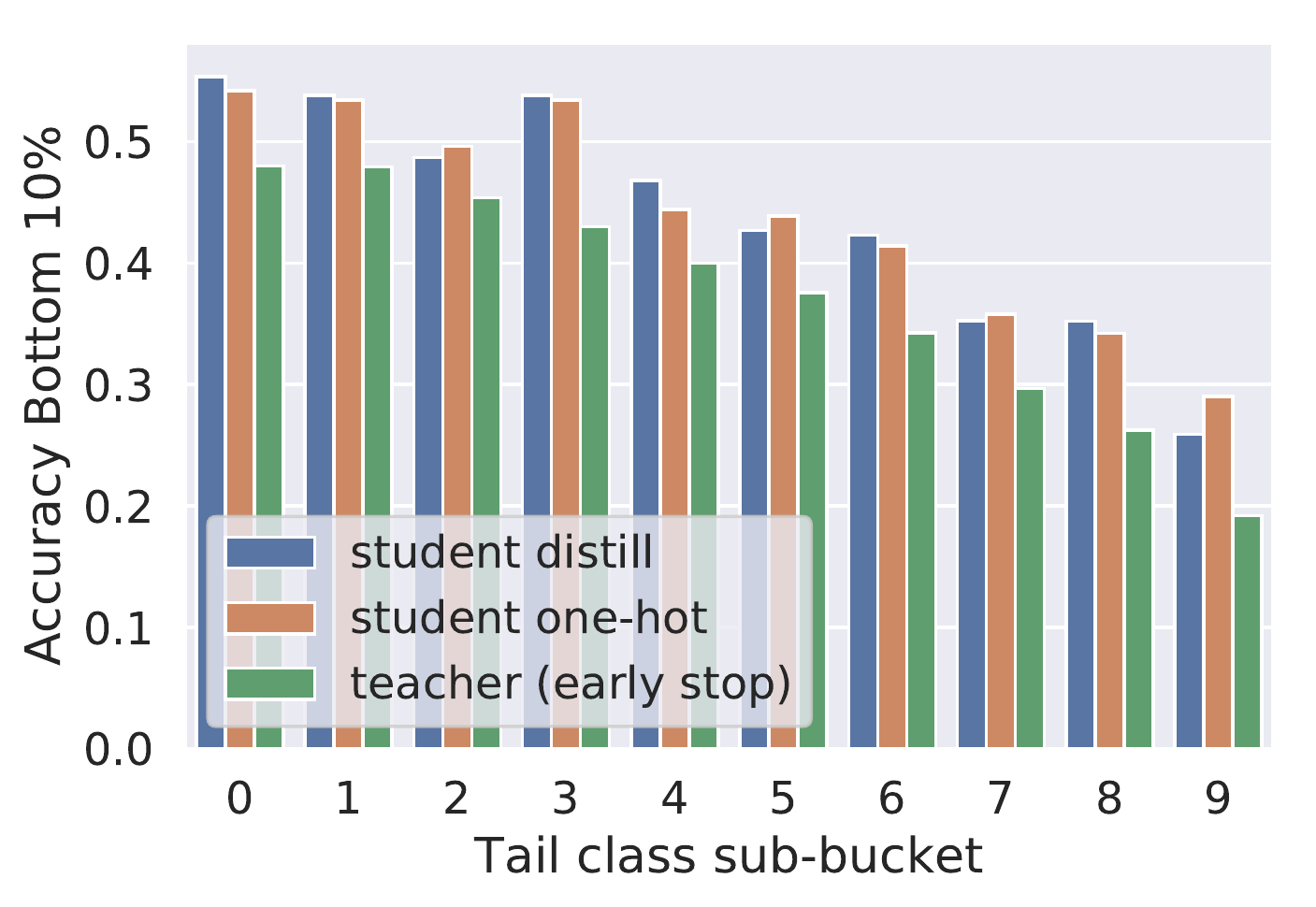}%
    }    
    \hfill
    \subfigure[Log-loss (tail classes).]{
    \includegraphics[scale=0.35]{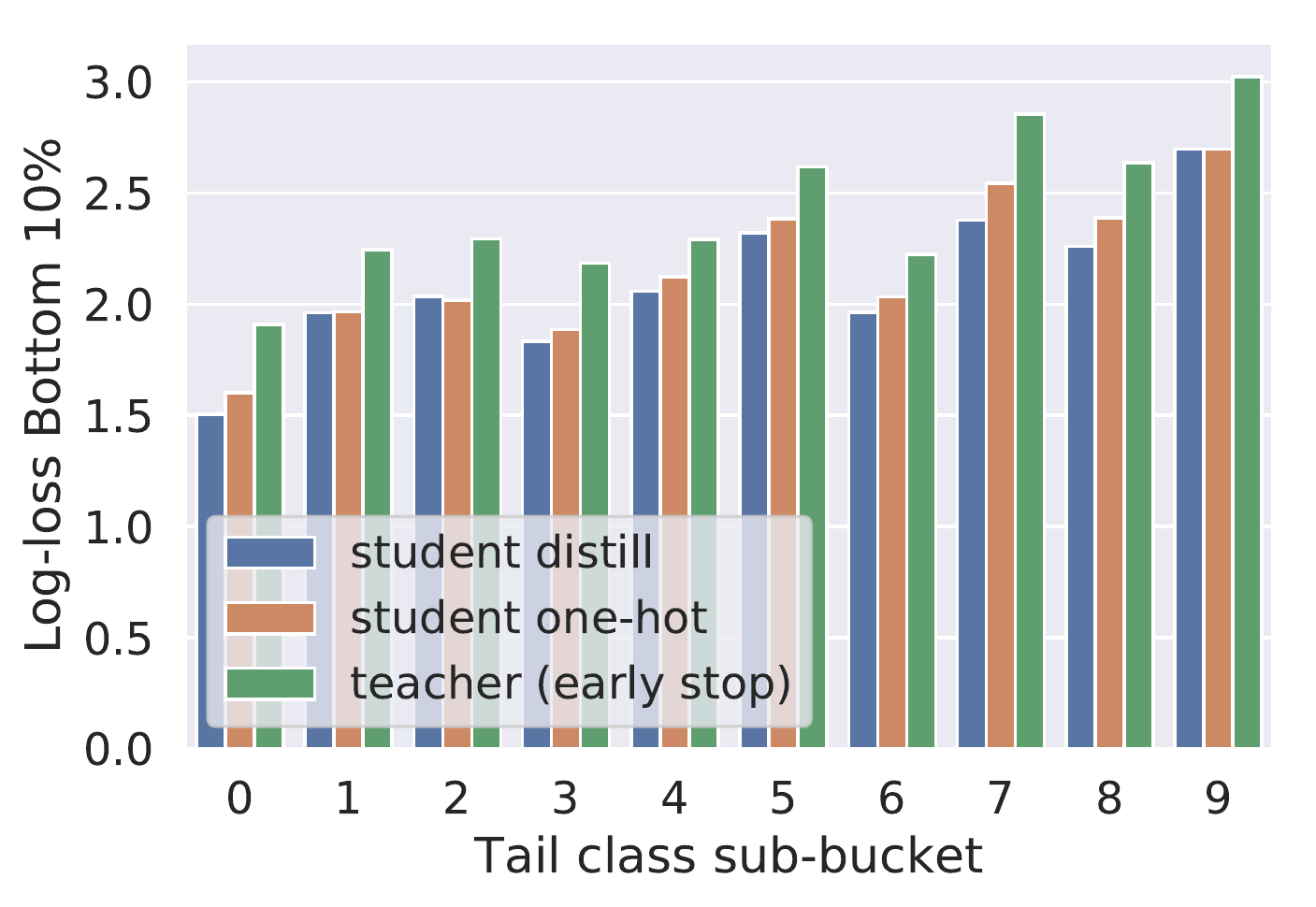}%
    }
    \hfill
    \subfigure[Margin (tail classes).]{
    \includegraphics[scale=0.35]{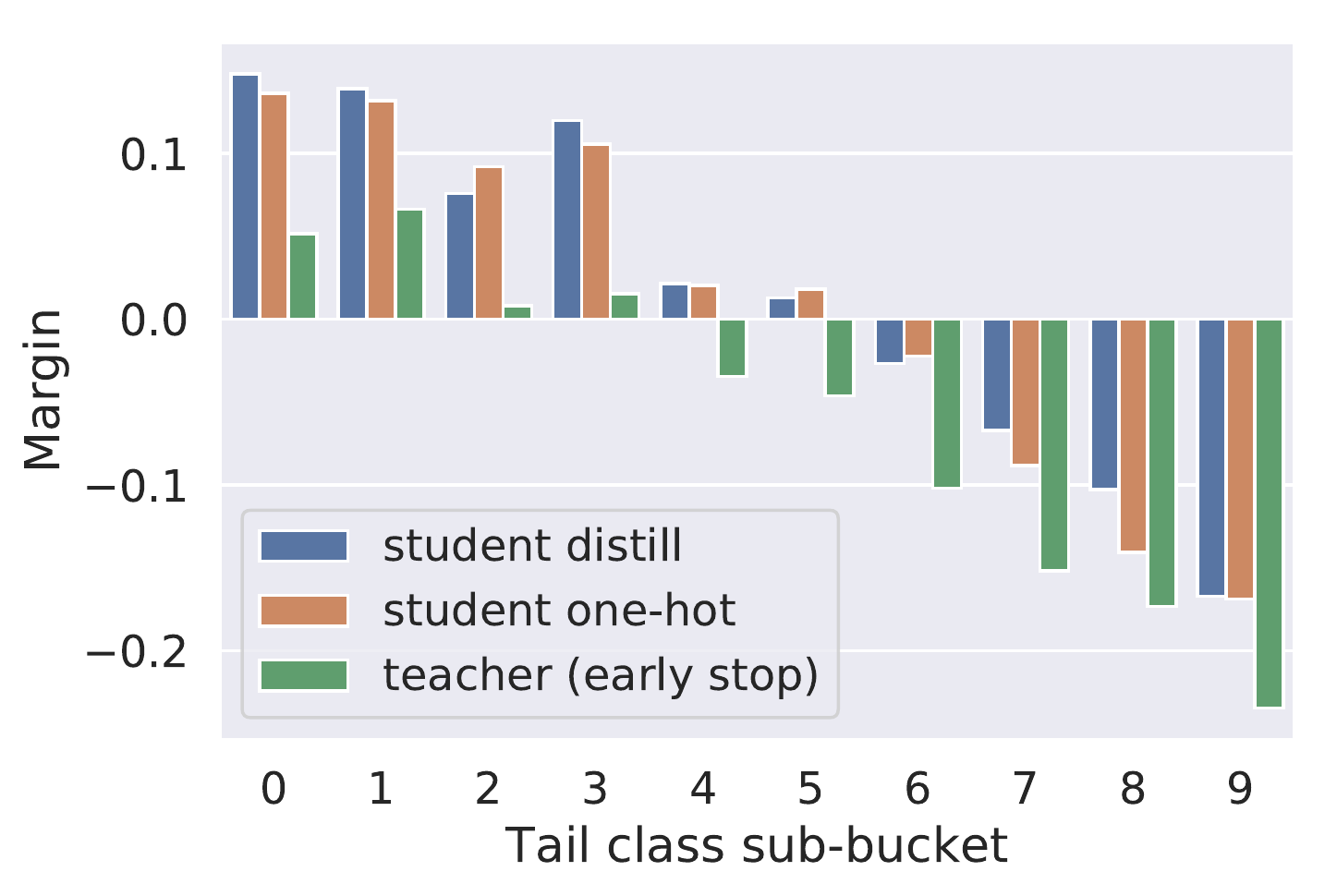}%
    }
    }

    \caption{Logit statistics
    for ResNet-50 self-distillation
    on
    ImageNet,
    for
the 
(early-stopped) teacher, 
self-distilled student, and one-hot (non-distilled) student.
Per Figure~\ref{fig:fair_dist_cifar100_lt_logit_breakdown},
we first create $10$ class buckets.
We zoom in on the ``tail'' bucket (comprising the $100$ ``hardest'' classes), and further split them into $10$ ``tail sub-buckets''.
As in Figure~\ref{fig:fair_dist_cifar100_lt_logit_breakdown}, the teacher is seen to confidently mispredict most samples on the last few buckets, with such misplaced confidence being transferred to the student.
    }
    \label{fig:fair_dist_imagenet_logit_breakdown}
\end{figure*}

On Imagenet,
we summarise statistics for \emph{three} models:
the early-stopped teacher, distilled student, and the one-hot (non-distilled) student.
As with CIFAR-100-LT, we sort classes by teacher accuracy, and bucket them into $10$ groups.
Owing to the larger number of labels,
we further zoom into the ``tail'' bucket (comprising the $100$ ``hardest'' classes),
and split them into $10$ sub-buckets.
From Figure~\ref{fig:fair_dist_imagenet_logit_breakdown}, 
the distilled student performs worse than its one-hot counterpart on the last bucket; this is in keeping with our results in Table~\ref{tbl:dataset_summary}.

\begin{figure*}[!t]
    \centering
    \includegraphics[scale=0.35]{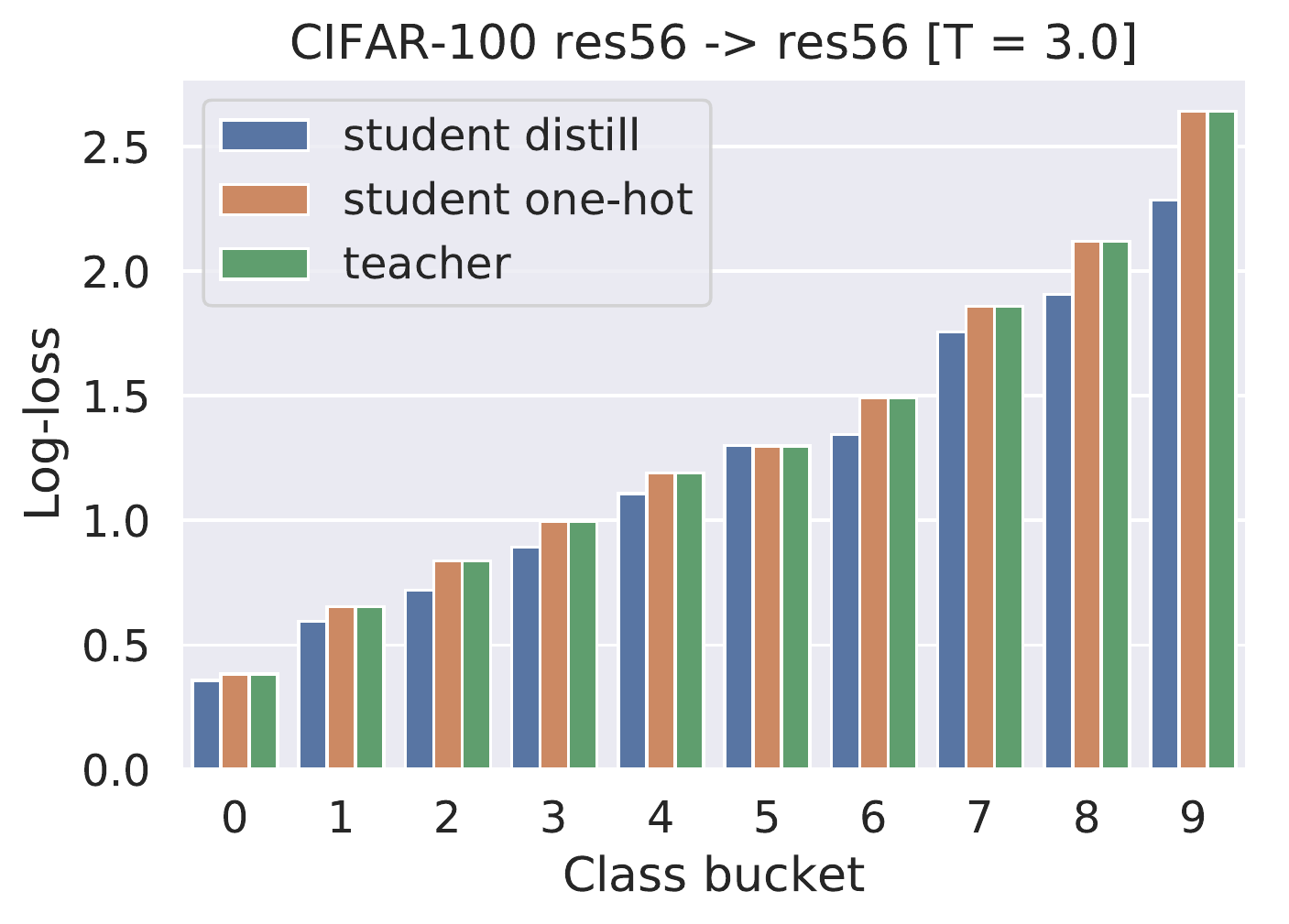}%
    \includegraphics[scale=0.35]{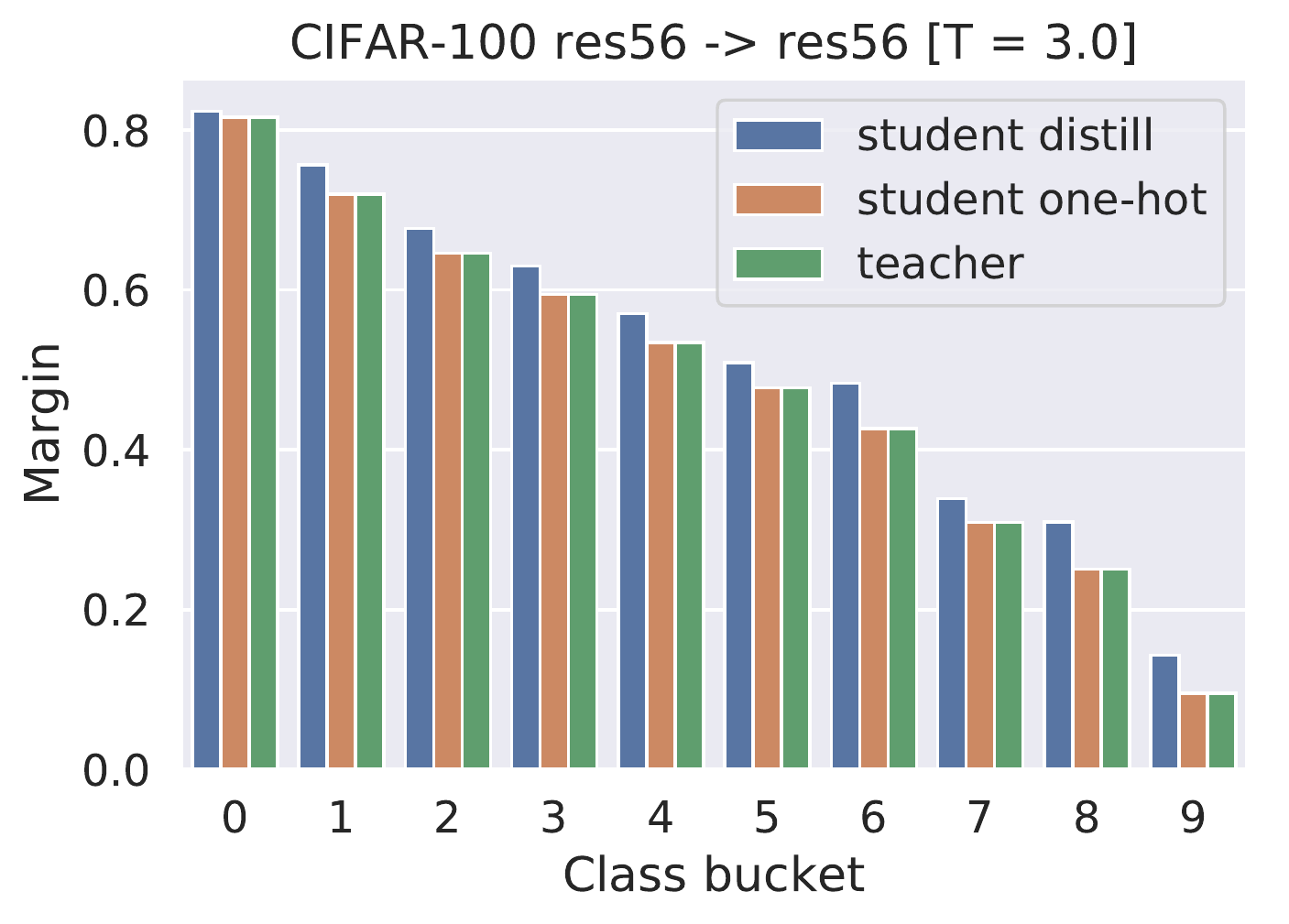}%
    
    \includegraphics[scale=0.35]{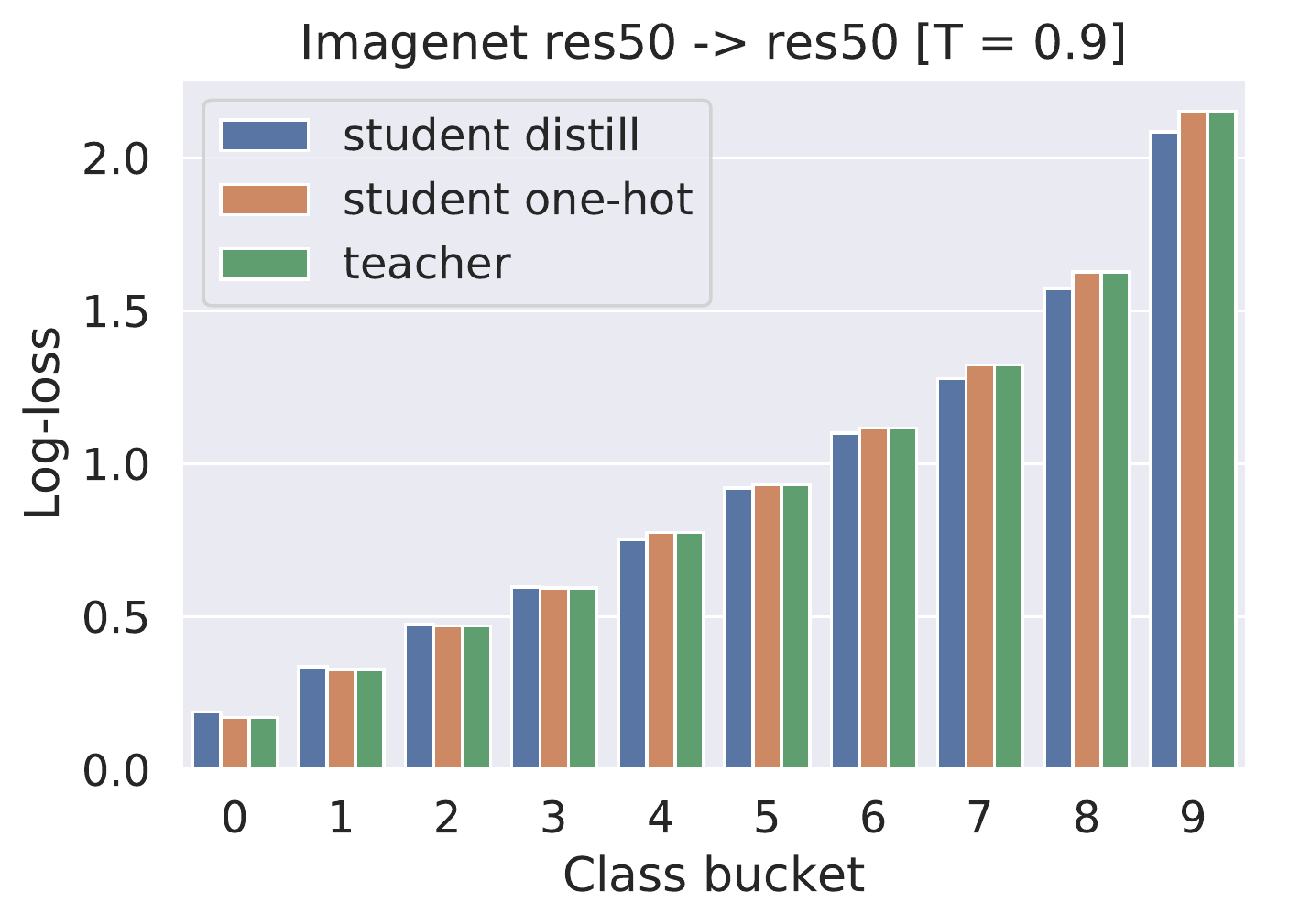}
    \includegraphics[scale=0.35]{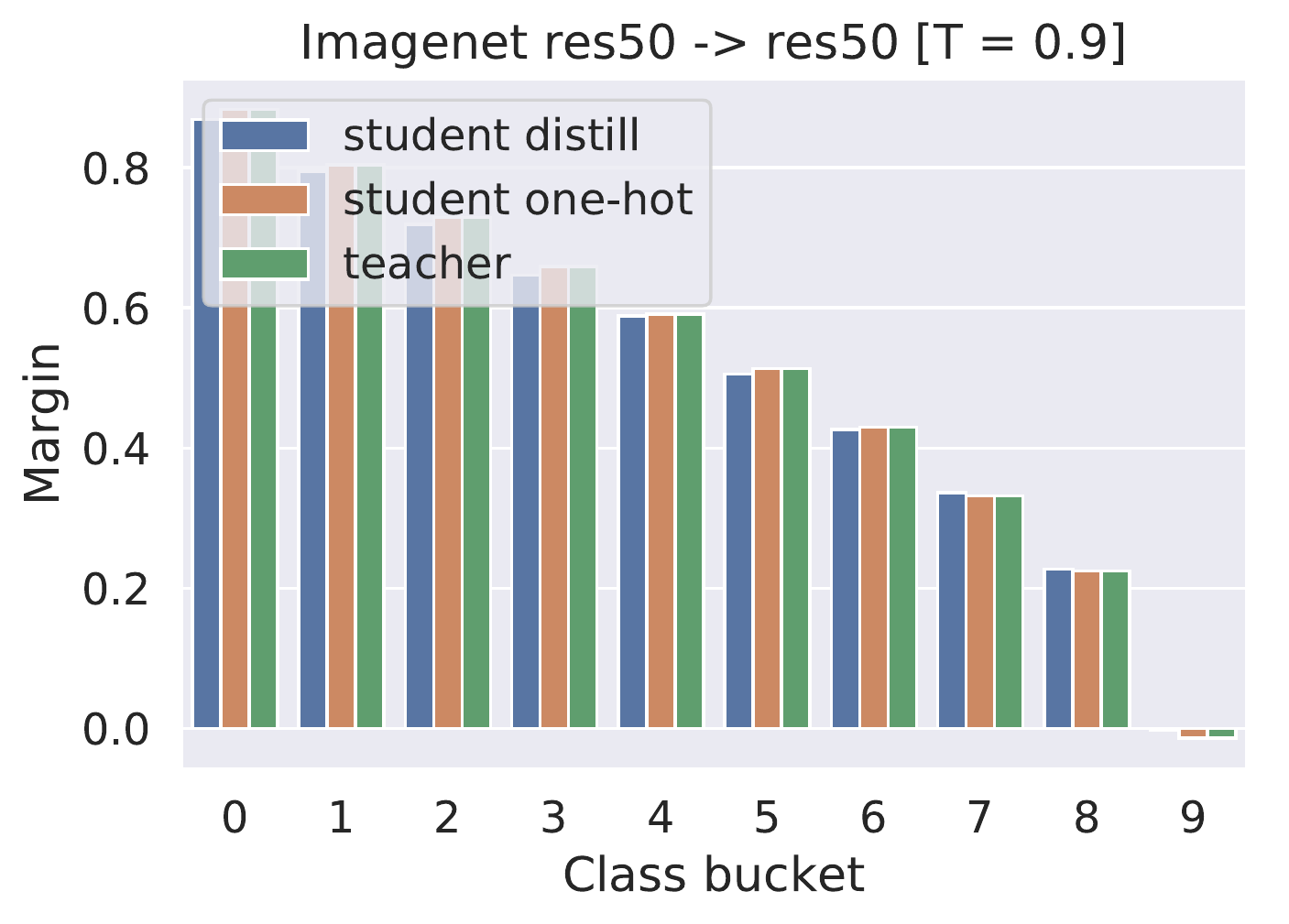}%

    \includegraphics[scale=0.35]{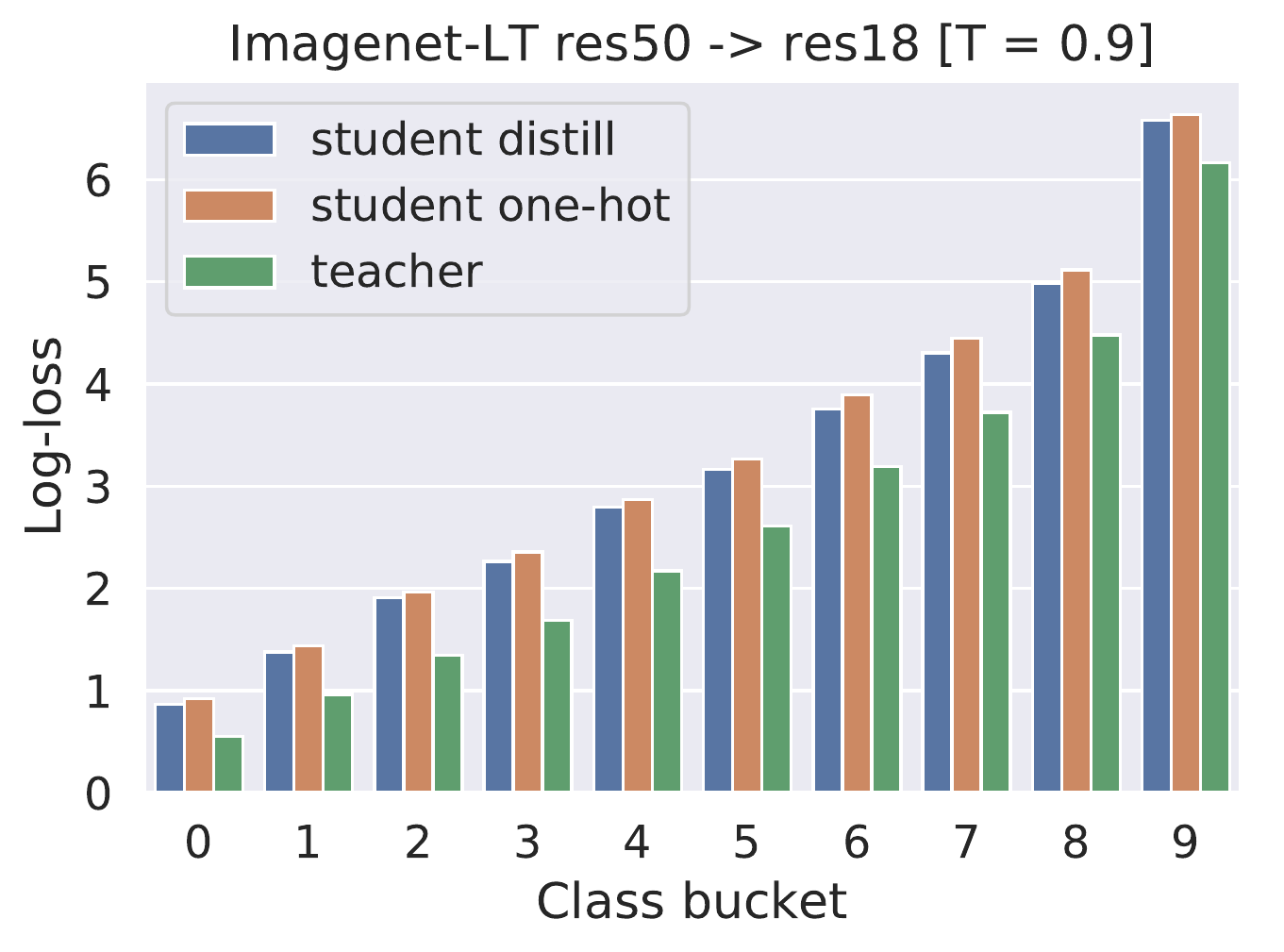}%
    \includegraphics[scale=0.35]{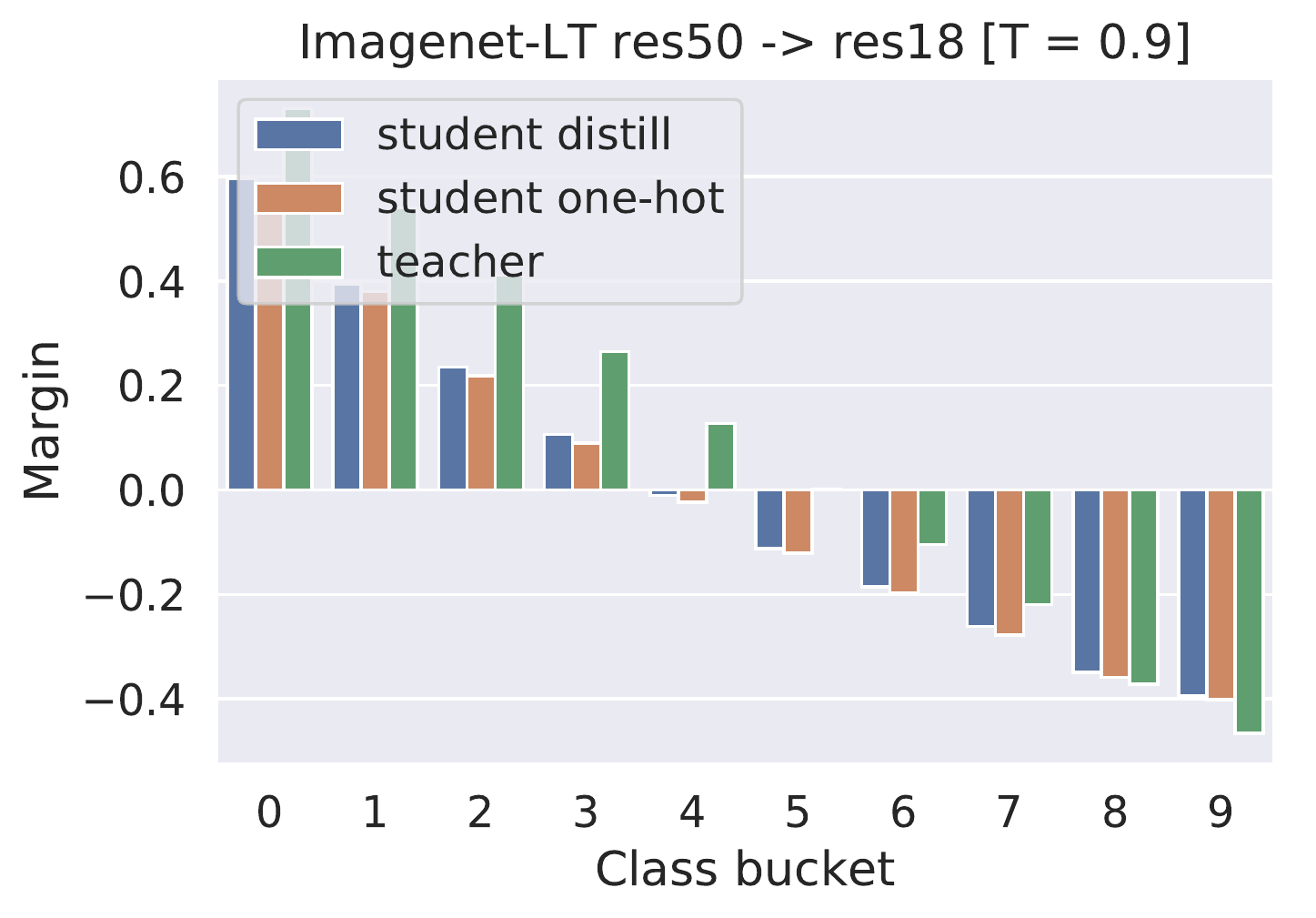}

    \caption{Logit statistics
for the teacher, student with one-hot labels,
and student with distilled labels across: datasets and architectures.
    }
    \label{fig:fair_dist_datasets_logit_full_breakdown}
\end{figure*}

Figure~\ref{fig:fair_dist_datasets_logit_full_breakdown} shows logit statistics for additional settings to considered in the body.
On ImageNet-LT, e.g., we see again that the margin of the teacher model systematically worsens and becomes negative on the hardest classes.

In Table~\ref{tbl:inat} we report results from the inherently long-tailed iNaturalist 2018 dataset~\citep{VanHorn:2017}.
Our observations made for other considered datasets hold: adaptive margin method improves over both one hot and plain distillation in terms of the worst class accuracy. 
We also observe, how the average accuracy improves.

\begin{table*}[!th]
    \centering
    
    \renewcommand{\arraystretch}{1.25}
    \begin{tabular}{@{}lllll@{}}
        \toprule
        \multirow{2}{*}{\textbf{Method}} &
        \multicolumn{4}{c}{\textbf{Per-class accuracy statistics}}\\
        &
        \textbf{Mean} & \textbf{Worst 20} & \textbf{Top 20\%} & \textbf{$\Delta$20}\\
\toprule
One-hot & $53.00$ & $5.00$ & $100.00$ & $48.00$\\
Distill & $52.67$ & $5.00$ & $100.00$ & $47.67$\\
AdaMargin & $53.33$ & $8.33$ & $98.33$ & $45.00$\\
AdaAlpha avg & $54.67$ & $13.33$ & $100.00$ & $41.33$\\
  \bottomrule
\end{tabular}

    \caption{Self-distillation experiments (from Resnet-50 to Resnet-50) on the iNaturalist dataset \cite{iNaturalist} with student's average accuracy using one-hot and distilled labels.
    Worst 20 denotes accuracy averaged over worst 20 classes
    $\Delta$20 denotes the difference between the mean accuracy and the worst 20 classified classes. 
    The proposed AdaMargin technique improves mean and worst-class  accuracy over both one-hot training and standard distillation.}
    \label{tbl:inat}
\end{table*}

\subsection{Logit plots under Ada-* methods}

\begin{figure*}[!t]
    \centering
    \subfigure[Accuracy.]{
    \includegraphics[scale=0.3]{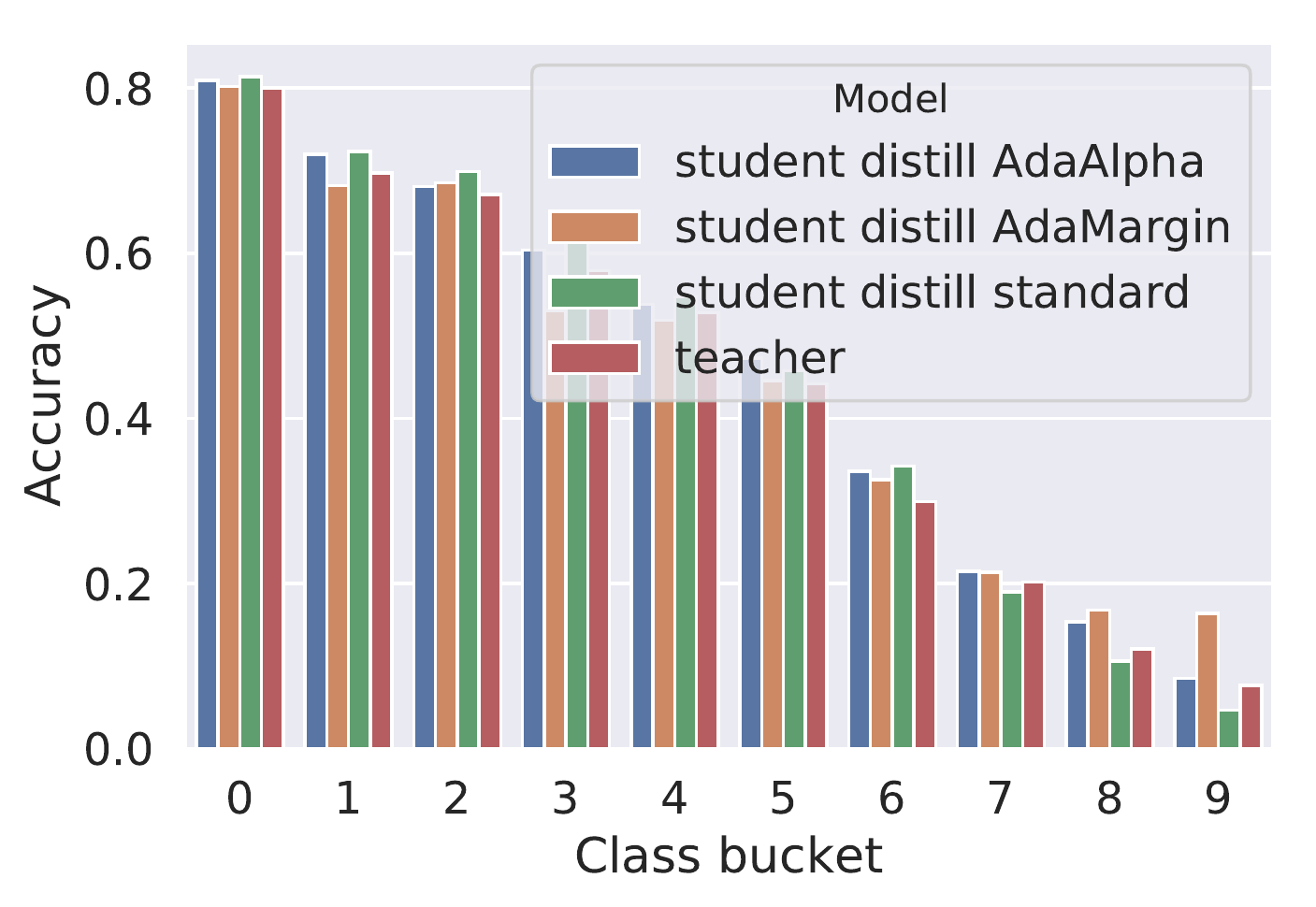}%
    }    
    \hfill
    \subfigure[Log-loss.]{
    \includegraphics[scale=0.3]{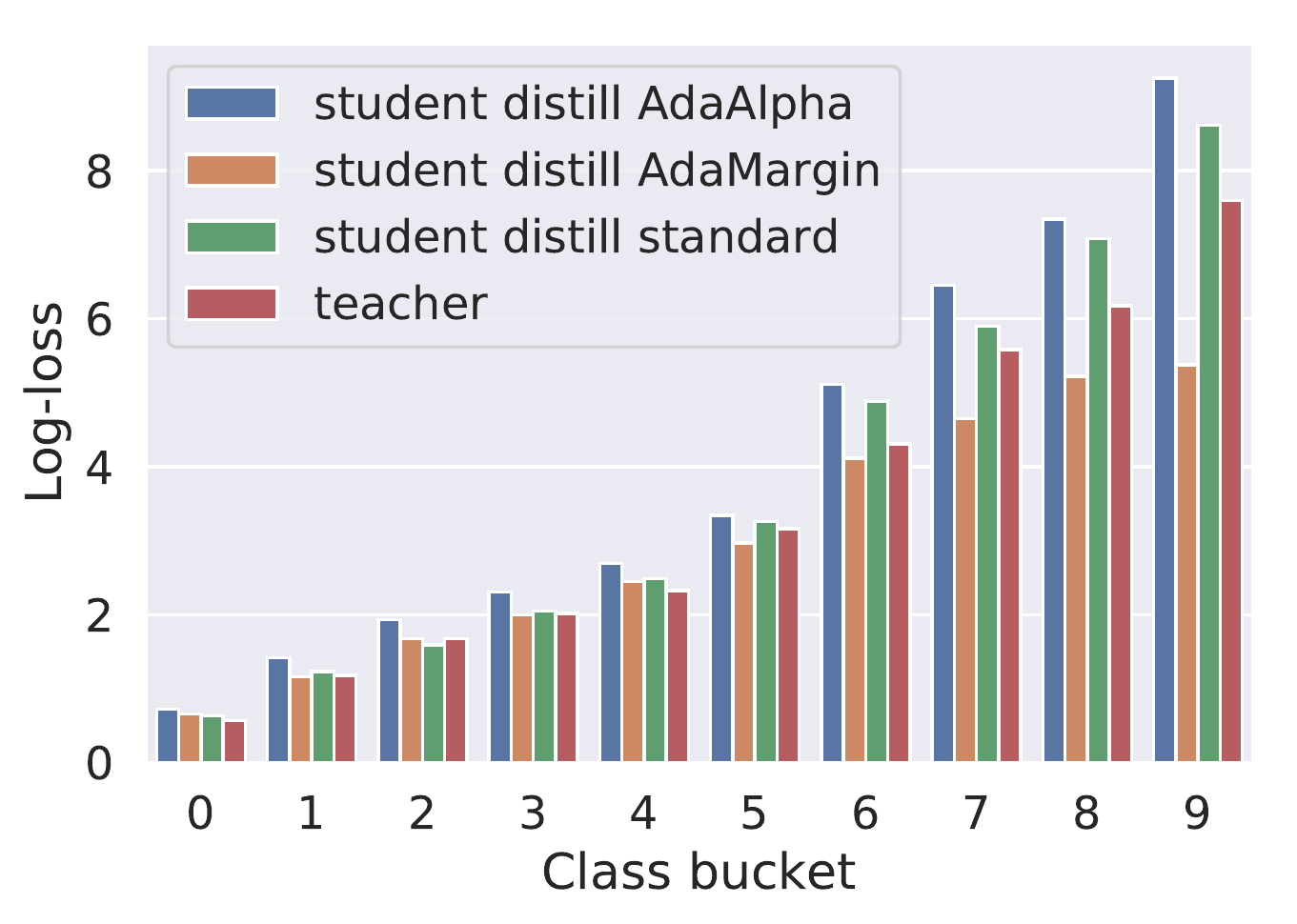}%
    }
    \hfill
    \subfigure[Margin).]{
    \includegraphics[scale=0.3]{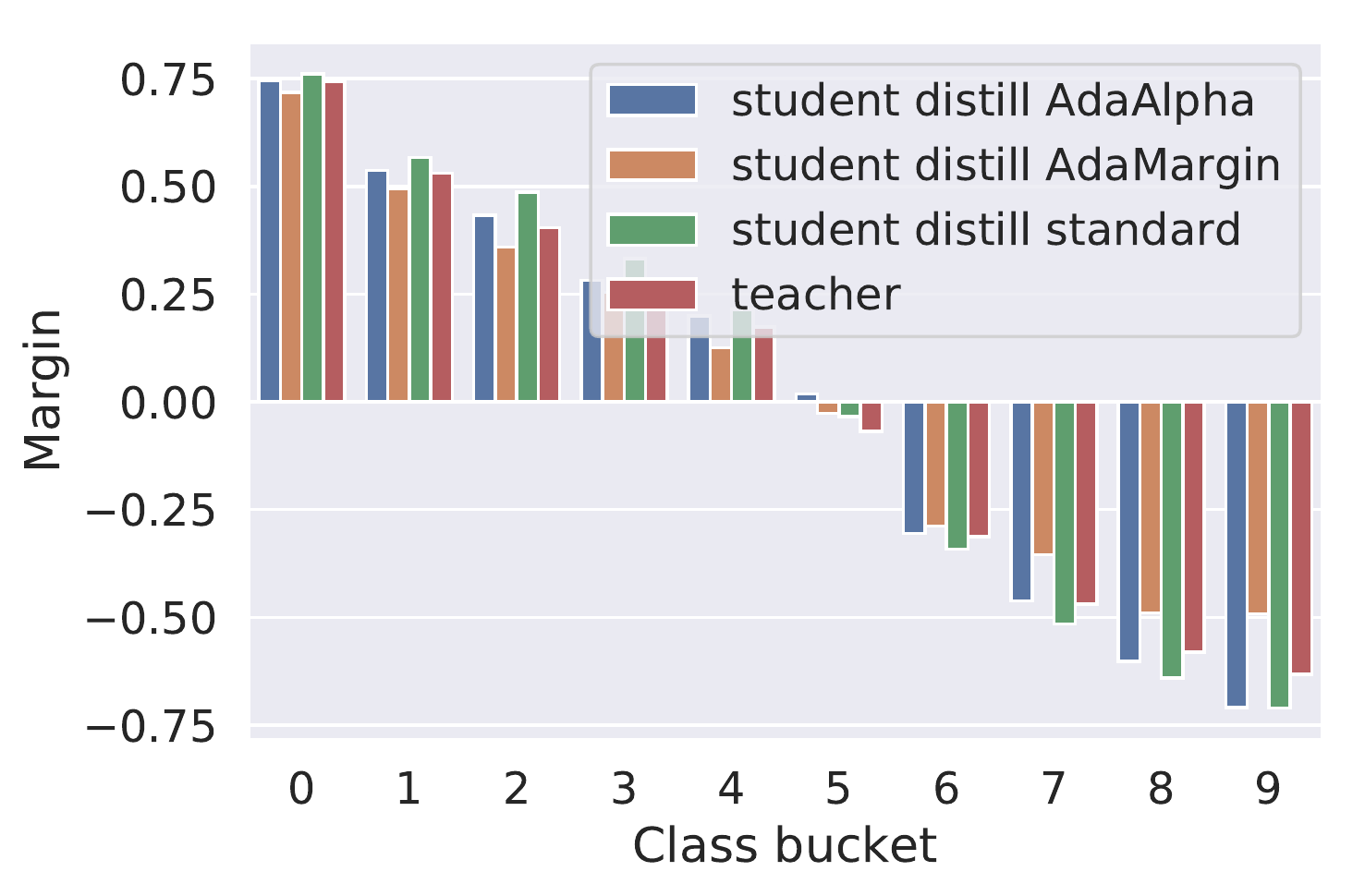}%
    }

    \caption{Logit statistics
    for ResNet-56 self-distillation
    on
    CIFAR-100 LT,
    for
the teacher, 
self-distilled student, and our adaptive methods.
Per Figure~\ref{fig:fair_dist_cifar100_lt_logit_breakdown},
we create $10$ class buckets.
AdaMargin flattens both the margin and log-loss distributions. AdaAlpha increases log loss across classes, while improving margins on few buckets, including flipping the bucket \emph{5} to have positive margin.
    }
    \label{fig:fair_dist_cifar100lt_adamethods_logit_breakdown}
\end{figure*}

Figure~\ref{fig:fair_dist_cifar100lt_adamethods_logit_breakdown}
shows the logit statistics under the proposed AdaMargin and AdaAlpha methods on CIFAR-100 LT.
We see that AdaMargin can generally improve the student margin and accuracy on the hardest classes, while also reducing the log-loss.
This confirms that the gains of the method come from improving behaviour of the scores on these hard classes.
We discuss this Figure in more detail in Section~\ref{s:results_and_discussion}.

\subsection{Results on Adult dataset}
\label{app:adult}

\newedit{We report the results of an experiment on the UCI Adult dataset.
This data comprises $\sim 48K$ examples, with the target being a binary label denoting whether or not an individual has income $\geq 50$K.
The data is mildly imbalanced, with $24\%$ of samples being positive.

Inspired by~\citet{Dao:2021}, we consider a random forest based distillation setup:
we use a teacher model that is a random forest \emph{classifier} comprising $500$ trees with a maximum depth of $20$,
and a student model that is a random forest \emph{regressor} comprising $1$ tree with a maximum depth of $20$.
The teacher model achieves a test (balanced) accuracy of $81.8\%$.

We perform distillation by feeding the student model the teacher's prediction scores, mixed in with the binary training labels with a weight $\alpha = 0.9$.
Distillation improves the student's overall (balanced) accuracy
significantly, from $76.2\%$ to $79.3\%$.
However, this gain is not distributed uniformly:
using per-label subgroups, we find that distillation helps the positive class by $+7.4\%$, but hurts the negative class by $-1.2\%$.
While by itself suggestive of asymmetry in distillation performance,
the data admits an arguably more natural subgroup creation,
based on available sex and sex features.
For example, we find that amongst low-income males, distillation hurts by $-2.2\%$;
further restricting to those who are Asian Pacific-Islander, the degradation is $-5.9\%$.
This confirms that in scenarios where fairness may be a consideration,
a na\"{i}ve application of distillation may be inadmissible.
}

\begin{table}[!t]
    \centering
    
    \caption{Difference between distillation and one-hot performance on Adult dataset.
    Here, subgroups are defined by the sex and label.
    $\Delta$ refers to the difference between the distilled and one-hot student's accuracy on the subgroup.}
    \label{tbl:adult_sex}
    \vspace{2mm}
    
    \begin{tabular}{lrr}
    \toprule
      sex &  label &  $\Delta$ \\
    \midrule
        Male &      0 & -2.222 \\
      Female &      0 &  0.393 \\
      Female &      1 &  2.373 \\
        Male &      1 &  8.384 \\
    \bottomrule
    \end{tabular}
\end{table}

\begin{table}[!t]
    \centering
    
    \caption{Difference between distillation and one-hot performance on Adult dataset.
    Here, subgroups are defined by the sex, race, and label.
    $\Delta$ refers to the difference between the distilled and one-hot student's accuracy on the subgroup.}
    \label{tbl:adult_race_sex} 
    \vspace{2mm}
    
    \begin{tabular}{llrr}
    \toprule
                    race &   sex &  label &   $\Delta$ \\
    \midrule
      Amer-Indian-Eskimo &   Female &      1 & -66.667 \\
      Asian-Pac-Islander &     Male &      0 &  -5.941 \\
                   Other &     Male &      0 &  -4.347 \\
                   Black &   Female &      1 &  -2.381 \\
                   White &     Male &      0 &  -2.248 \\
                   Black &     Male &      0 &  -1.639 \\
                   Black &   Female &      0 &  -0.140 \\
      Asian-Pac-Islander &   Female &      0 &   0.000 \\
                   White &   Female &      0 &   0.388 \\
                   Other &   Female &      0 &   2.439 \\
                   White &   Female &      1 &   2.724 \\
      Amer-Indian-Eskimo &   Female &      0 &   6.349 \\
      Amer-Indian-Eskimo &     Male &      0 &   6.493 \\
      Asian-Pac-Islander &   Female &      1 &   7.692 \\
                   White &     Male &      1 &   7.796 \\
                   Black &     Male &      1 &   9.489 \\
                   Other &     Male &      1 &  15.000 \\
      Asian-Pac-Islander &     Male &      1 &  19.626 \\
                   Other &   Female &      1 &  20.000 \\
      Amer-Indian-Eskimo &     Male &      1 &  25.000 \\
    \bottomrule
    \end{tabular}
\end{table}

\subsection{Analysis of regularisation samples}
\label{app:regularisation_samples}

\begin{figure}[!t]
    \centering
    \subfigure[Distribution of regularisation samples per label bucket.]{%
    \includegraphics[scale=0.2]{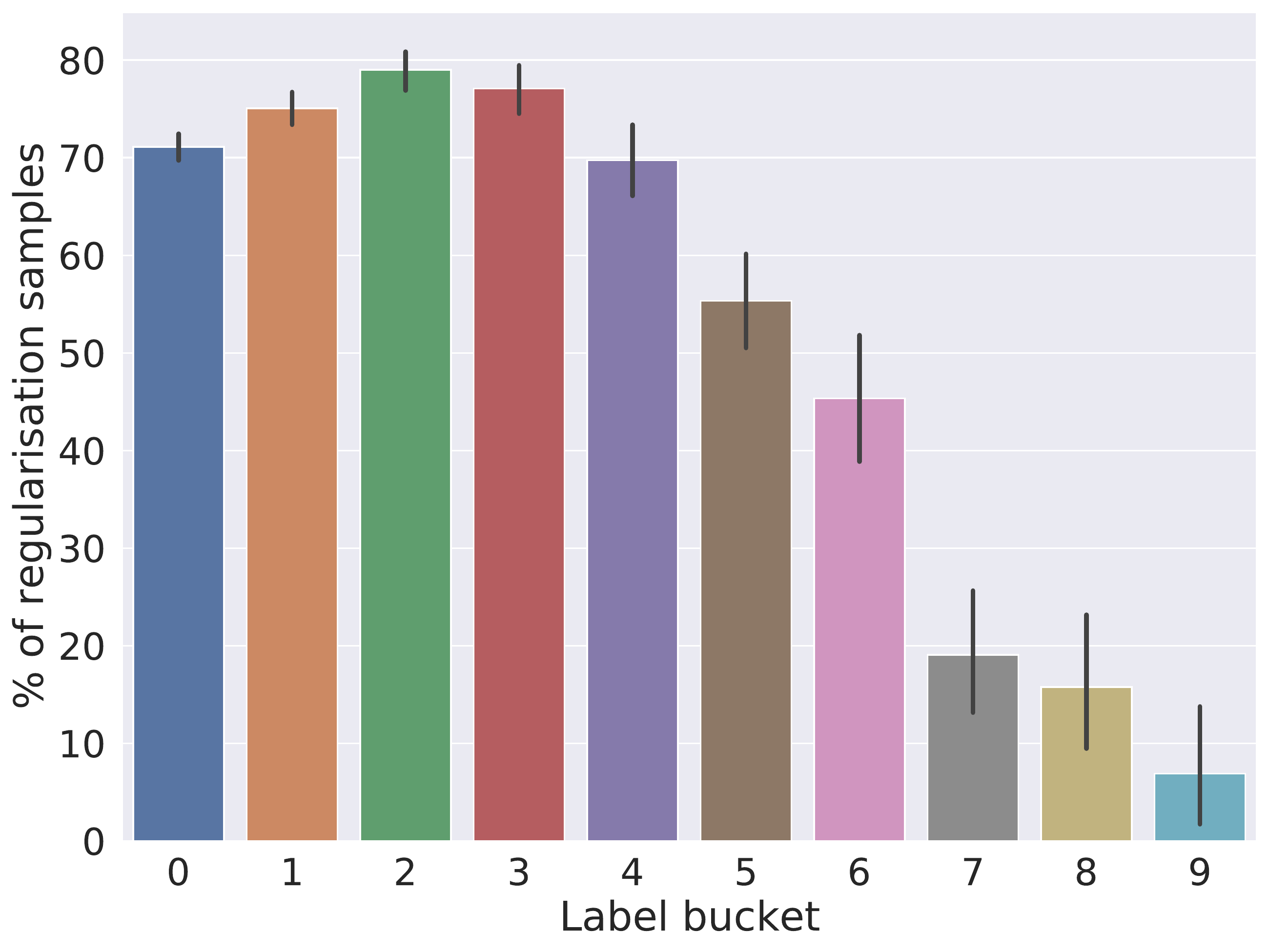}
    \label{fig:reg_samples}}
    \quad
    \subfigure[Teacher versus distilled student probailities on training label.]{%
    \includegraphics[scale=0.2]{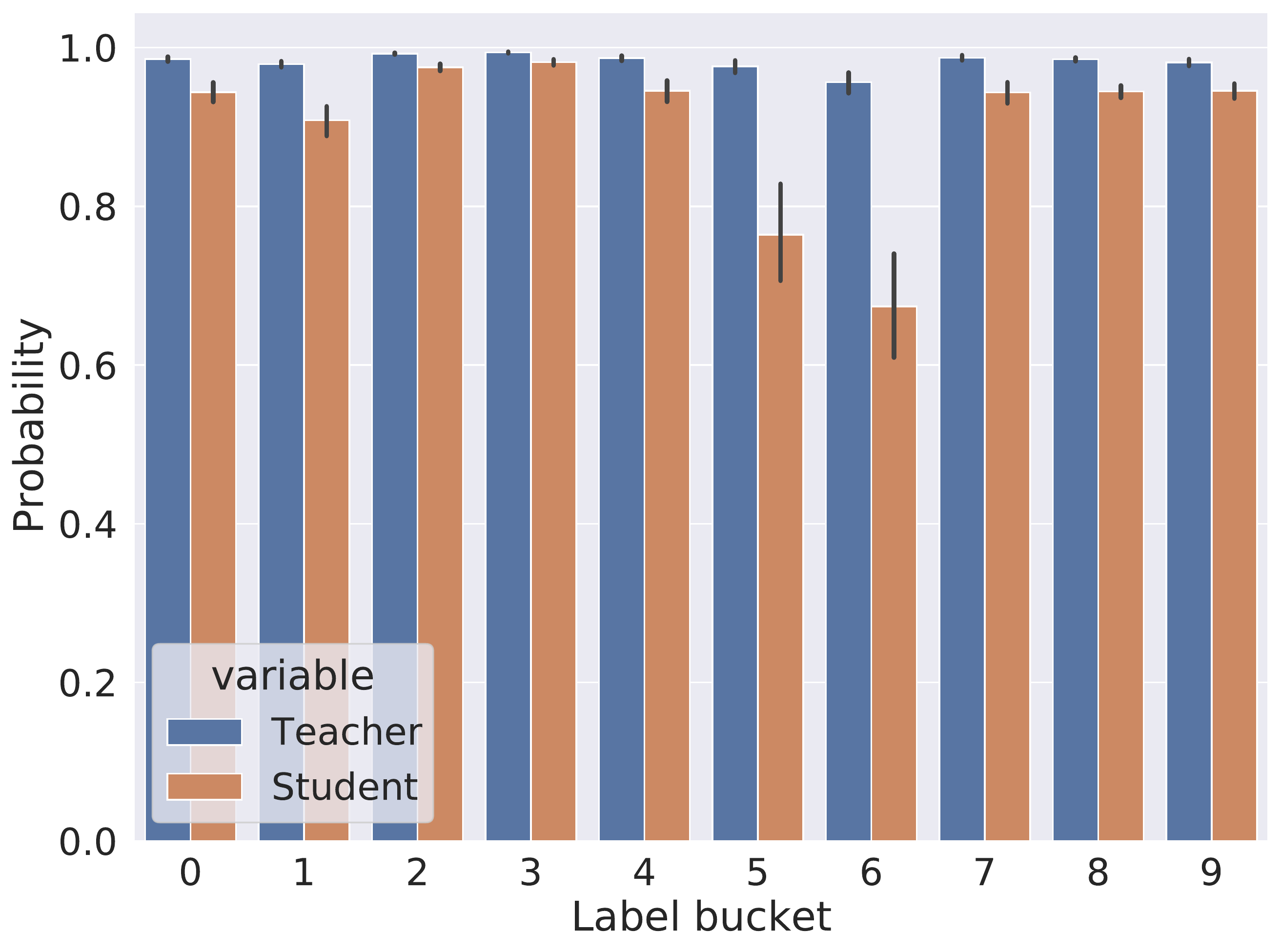}
    \label{fig:training_logits}}
    
    \caption{Study of regularisation samples on training set, CIFAR-100 LT.}
\end{figure}

\begin{figure}[!t]
    \centering
    \subfigure[Distribution of regularisation samples per label bucket.]{%
    \includegraphics[scale=0.2]{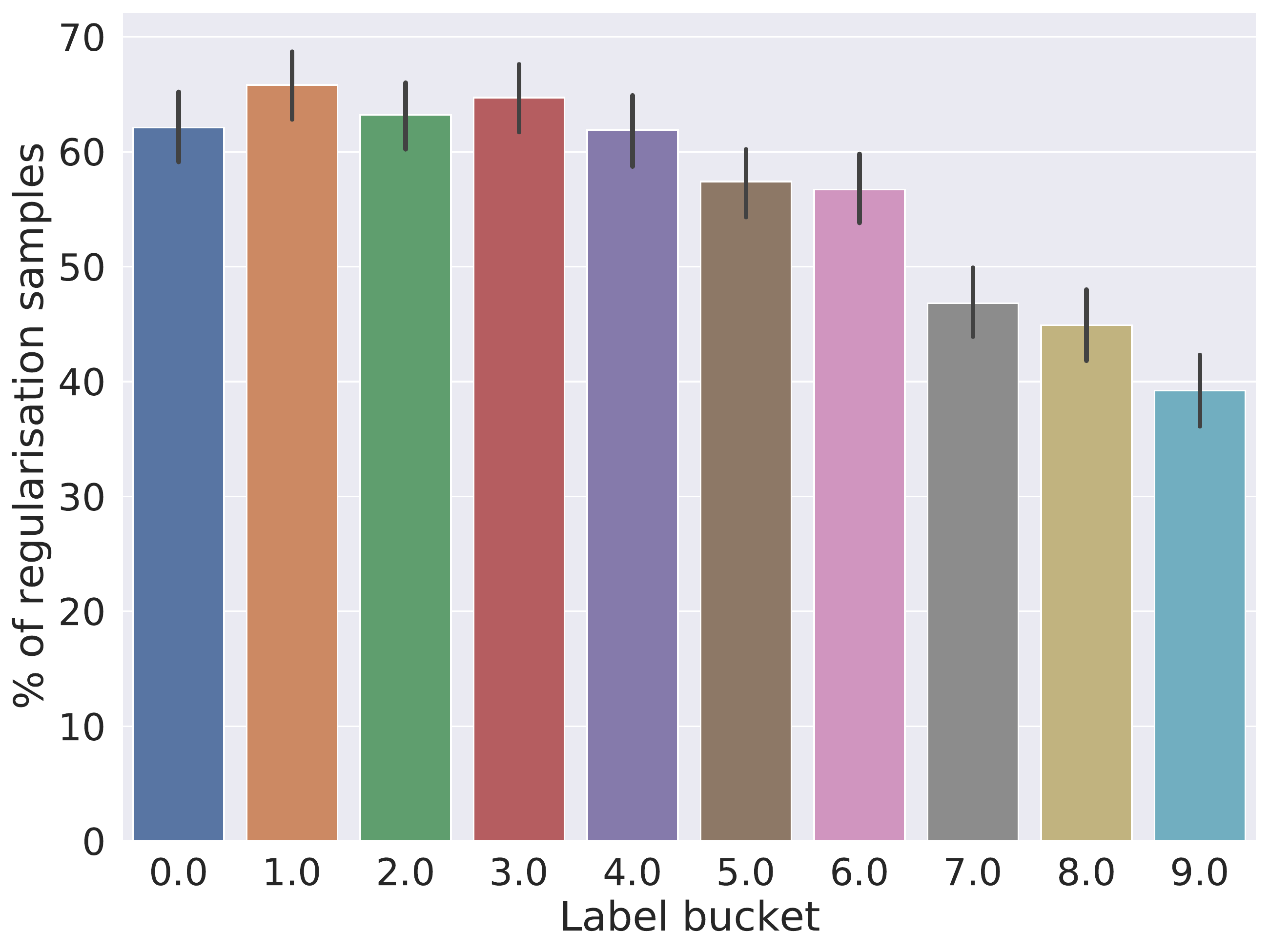}
    \label{fig:reg_samples_test}}
    \quad
    \subfigure[Teacher versus distilled student probailities on test label.]{%
    \includegraphics[scale=0.2]{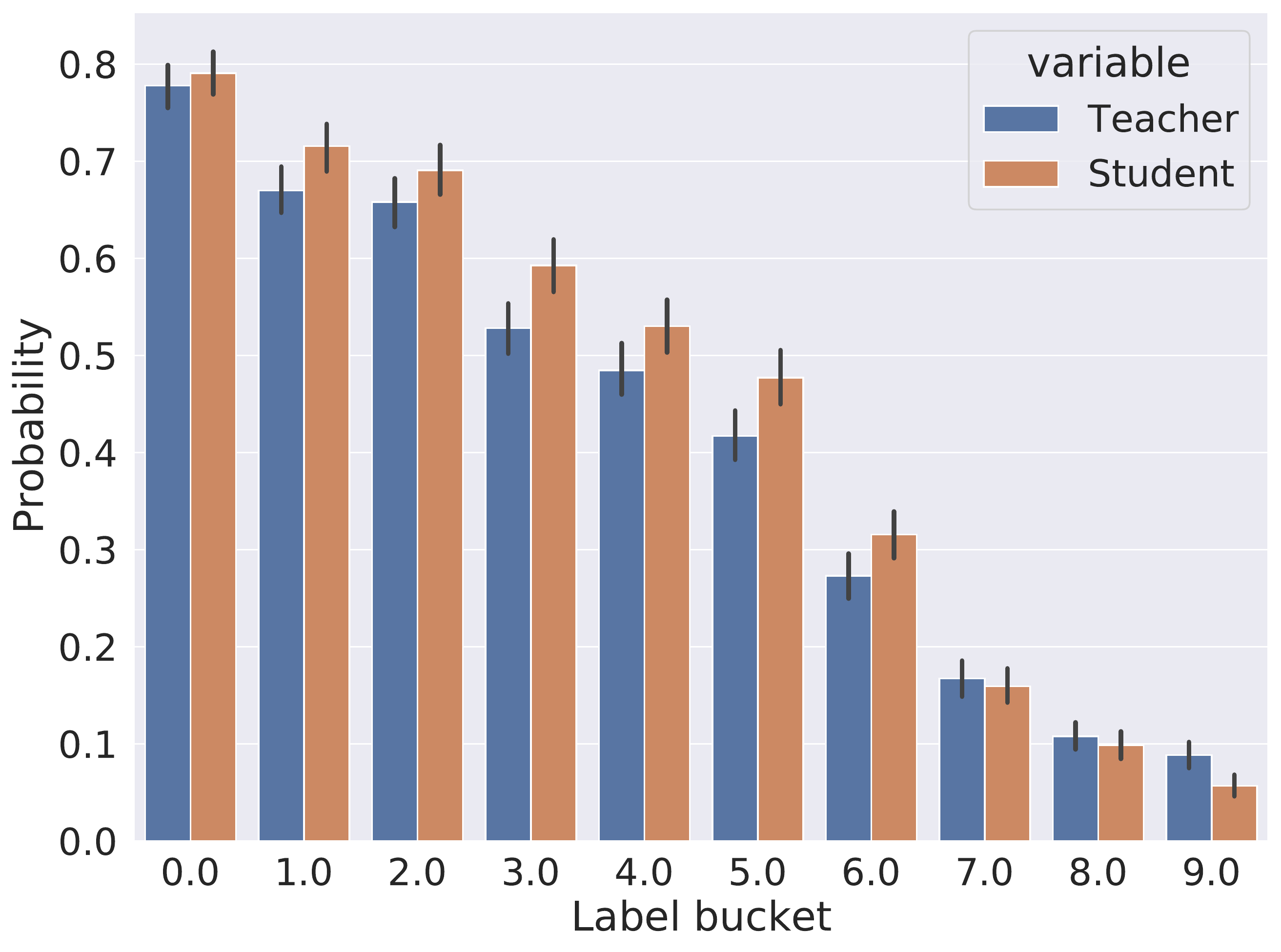}
    \label{fig:test_logits}}
    
    \caption{Study of regularisation samples on test set, CIFAR-100 LT.}
\end{figure}

\newedit{
Recently,~\citet{Zhou:2021} proposed the notion of \emph{regularisation samples} to understand how distillation's performance can be improved.
In brief, such samples correspond to cases where the teacher's prediction on the training label is less than the distilled student's prediction on this label;
these may be shown to correspond to cases where a certain notion of ``variance reduction'' dominates a notion of ``bias reduction''.
Given our analysis above of the asymmetric effects of distillation on certain subgroups, it is natural to consider whether or not these relate to the presence of regularisation samples in these groups.

Figure~\ref{fig:reg_samples} 
visualises the distribution of regularisation samples inside subgroups defined by $10$ label buckets.
where the labels are sorted in descending order of label frequency.
Here, we compare the predicted probabilities of the teacher and final distilled student models on all \emph{training} samples (as was done in the analysis of~\citet{Zhou:2021}).
Interestingly, we see that the tail buckets tend to have very few regularisation samples;
i.e., for rare labels, the teacher prediction on the training label is generally higher than that of the distilled student model.
We confirm this in Figure~\ref{fig:training_logits}.

While the analysis of~\citet{Zhou:2021} was primarily for training samples
---
since the aim in identifying regularisation samples was to mitigate their influence during training
---
we may also identify the breakdown of such samples on test data.
Figure~\ref{fig:reg_samples_test} shows that, compared to the training set,
there are in absolute terms more such samples across nearly every label bucket;
however, there is again no clear correlation between the label bucket and the fraction of such samples.
In particular, the tail bucket is again the one with the \emph{fewest} regularisation samples.
This is corroborated by the probability scores of the teacher and student in Figure~\ref{fig:test_logits}.

Overall, this results suggest that the existing notion of regularisation samples may not, by themselves,
be sufficient to predict the poor performance of distillation on certain subgroups defined by labels.
}

\subsection{Impact of repeated distillation}

In the body, we showed that performing distillation once can harm worst-class accuracy.
However, what is the effect of repeating this process, and distilling using the resulting student as a new teacher?
Does the worst-class accuracy get further harmed?

Table~\ref{tbl:repeated_dist} shows that on CIFAR-100 LT, repeating distillation can indeed harm worst-class performance,
even though  average performance remains roughly similar.
This further highlights the potential tension between average and worst-case performance under distillation.

\begin{table*}[!t]
    \centering
    
    \renewcommand{\arraystretch}{1.25}
    \begin{tabular}{@{}lllll@{}}
        \toprule \multirow{2}{*}{\textbf{Method}} &
        \multicolumn{3}{c}{\textbf{Per-class accuracy statistics}}\\
        &
        \textbf{Mean} & \textbf{Worst 10} & \textbf{Top 10\%} & \textbf{$\Delta$10}\\
\toprule
One-hot & $44.16$ & $3.00$ & $87.70$ & $41.16$ \\
Distillation 1$\times$ & $45.49$ & $0.90$ & $88.10$ & $44.59$ \\
Distillation 2$\times$  & $45.22$ & $0.00$ & $88.40$ & $45.22$ \\
Distillation 3$\times$  &$44.80$ & $0.00$ & $87.60$ & $44.80$ \\
  \bottomrule
\end{tabular}

    \caption{Results of repeated distillation on CIFAR-100 LT.
    Using a distilled student as teacher for a subsequent round of distillation is seen to further hurt worst-class accuracy.
    }
    \label{tbl:repeated_dist}
\end{table*}

\end{document}